\documentclass{article}

    \usepackage[preprint]{neurips_2025}

\usepackage{amssymb}
\usepackage{amsmath}
\usepackage[caption = true]{subfig}
\usepackage[utf8]{inputenc} 
\usepackage[T1]{fontenc}    
\usepackage{hyperref}       
\usepackage{booktabs}
\usepackage{url}            
\usepackage{booktabs}       
\usepackage{amsfonts}       
\usepackage{nicefrac}       
\usepackage{microtype}      
\usepackage{xcolor}         
\usepackage[ruled, lined, linesnumbered, commentsnumbered, longend]{algorithm2e}
\usepackage{graphicx}
\usepackage{float}
\usepackage{dirtytalk}
\usepackage{soul}
\usepackage{multirow}
\usepackage{colortbl}
\usepackage{comment}
\usepackage{bbm}

\newcommand{\pmval}[2]{#1 {\scriptsize{($\pm$ #2)}}}

\definecolor{temporal}{HTML}{fae0c7}  
\definecolor{spatio}{HTML}{deb5d7}  
\definecolor{proto}{HTML}{b5ddda}  

\usepackage{amsmath,amsfonts,bm}

\def\eqref#1{Eq.~(\ref{#1})}

\def\1{\bm{1}}

\def\rvg{{\mathbf{g}}}

\DeclareMathAlphabet{\mathsfit}{\encodingdefault}{\sfdefault}{m}{sl}
\SetMathAlphabet{\mathsfit}{bold}{\encodingdefault}{\sfdefault}{bx}{n}

\def\gD{{\mathcal{D}}}
\def\gE{{\mathcal{E}}}

\def\gL{{\mathcal{L}}}
\def\gM{{\mathcal{M}}}
\def\gN{{\mathcal{N}}}

\def\gS{{\mathcal{S}}}
\def\gT{{\mathcal{T}}}
\def\gU{{\mathcal{U}}}

\newcommand{\E}{\mathbb{E}}

\DeclareMathOperator*{\argmin}{arg\,min}

\newtheorem{definition}{Definition}
\newtheorem{theorem}{Theorem}

\newtheorem{example}{Example}

\title{Improving Generalization in Heterogeneous Federated Continual Learning via Spatio-Temporal Gradient Matching with Prototypical Coreset}

\author{
  Minh-Duong~Nguyen$^\heartsuit$, Le-Tuan Nguyen$^\heartsuit$, Quoc-Viet Pham$^\spadesuit$\\
  $^\heartsuit$Equal Contribution, \\
  $^\spadesuit$Trinity College Dublin, Ireland\\
  \texttt{\{mduongbkhn, letuanhf\}@gmail.com, viet.pham@tcd.ie} 
}

\begin{document}

\maketitle

\begin{abstract}
Federated Continual Learning (FCL) has recently emerged as a crucial research area, as data from distributed clients typically arrives as a stream, requiring sequential learning. This paper explores a more practical and challenging FCL setting, where clients may have unrelated or even conflicting data and tasks. In this scenario, statistical heterogeneity and data noise can create spurious correlations, leading to biased feature learning and catastrophic forgetting. Existing FCL approaches often use generative replay to create pseudo-datasets of previous tasks. However, generative replay itself suffers from catastrophic forgetting and task divergence among clients, leading to overfitting in FCL. To address these challenges, we propose a novel approach called \textbf{\underline{S}}patio-\textbf{\underline{T}}emporal gr\textbf{\underline{A}}dient \textbf{\underline{M}}atching with network-free \textbf{\underline{P}}rototype (STAMP). Our contributions are threefold: 1) We develop a model-agnostic method to determine subset of samples that effectively form prototypes when using a prototypical network, making it resilient to continual learning challenges; 2) We introduce a spatio-temporal gradient matching approach, applied at both the client-side (temporal) and server-side (spatial), to mitigate catastrophic forgetting and data heterogeneity; 3) We leverage prototypes to approximate task-wise gradients, improving gradient matching on the client-side. Extensive experiments demonstrate our method’s superiority over existing baselines.
\end{abstract}

\section{Introduction}
Federated Learning (FL) is a distributed and privacy-preserving approach that facilitates collaboration among various entities, such as organizations or devices \citep{9060868}. In FL, multiple clients train a shared model in coordination with a central server without exchanging personal data. Recently, FL has gained significant attention across multiple domains, including healthcare \citep{10.1145/3501296}, Internet-of-Things (IoT) \citep{9829327}, and autonomous driving \citep{9981098}. While conventional FL studies often assume static data classes and domains, real-world scenarios involve the continuous emergence of new classes and evolving data distributions \citep{elsayed2024addressing}. Training entirely new models to accommodate these changes is impractical due to the substantial computational resources required. An alternative approach is transfer learning from pre-trained models; however, this method is prone to catastrophic forgetting \citep{10323204}, which degrades performance on previously learned classes. To address catastrophic forgetting in FL, recent research \citep{Luo_2023_CVPR} has introduced the concept of Federated Continual Learning (FCL), which integrates the principles of FL and Continual Learning (CL) \citep{Yang_2023_ICCV}.

\begin{figure}[ht]
\vspace{-5mm}
\centering
\subfloat[Generalization gap on CIFAR100\label{fig:fedavg-cosine}]{\includegraphics[width=0.45\linewidth]{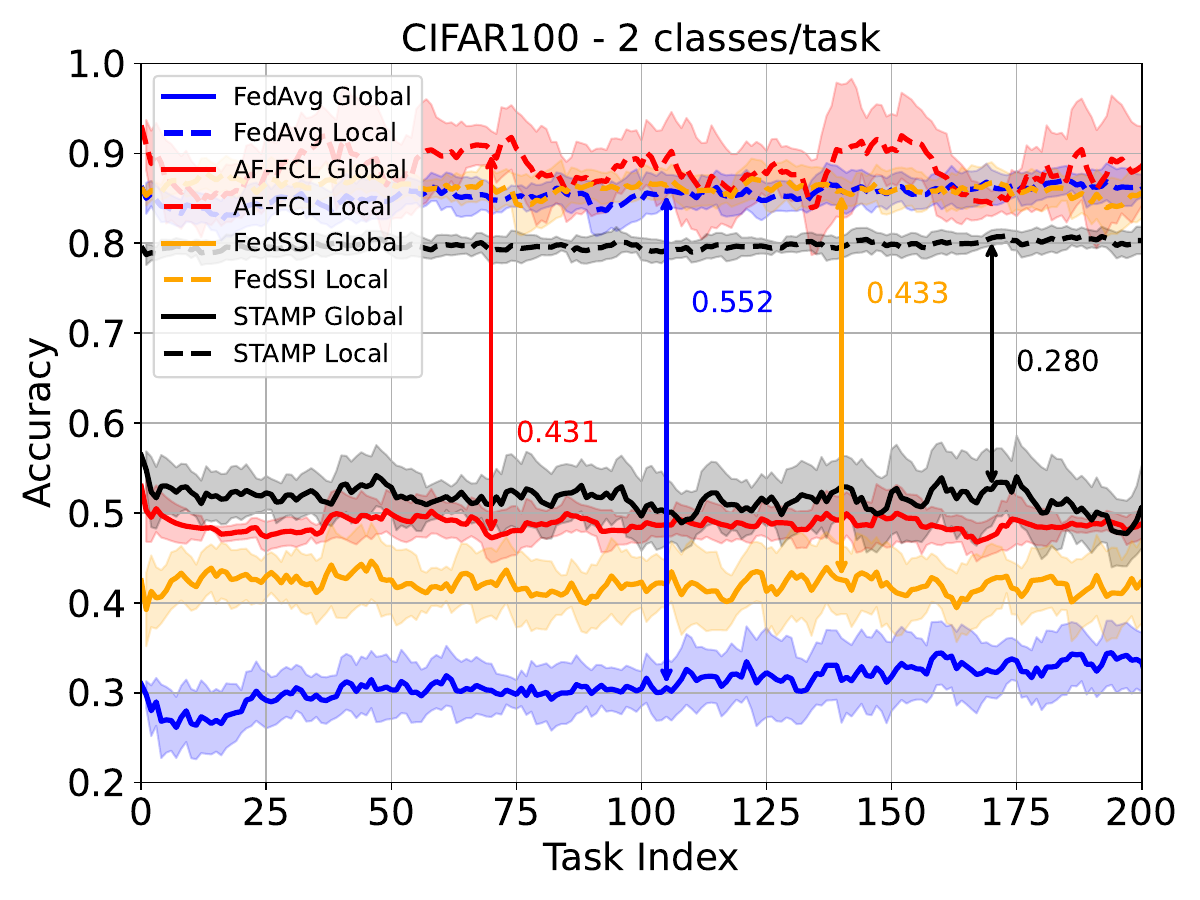}}
\subfloat[Generalization gap on ImageNet1K\label{fig:local-global}]{\includegraphics[width=0.45\linewidth]{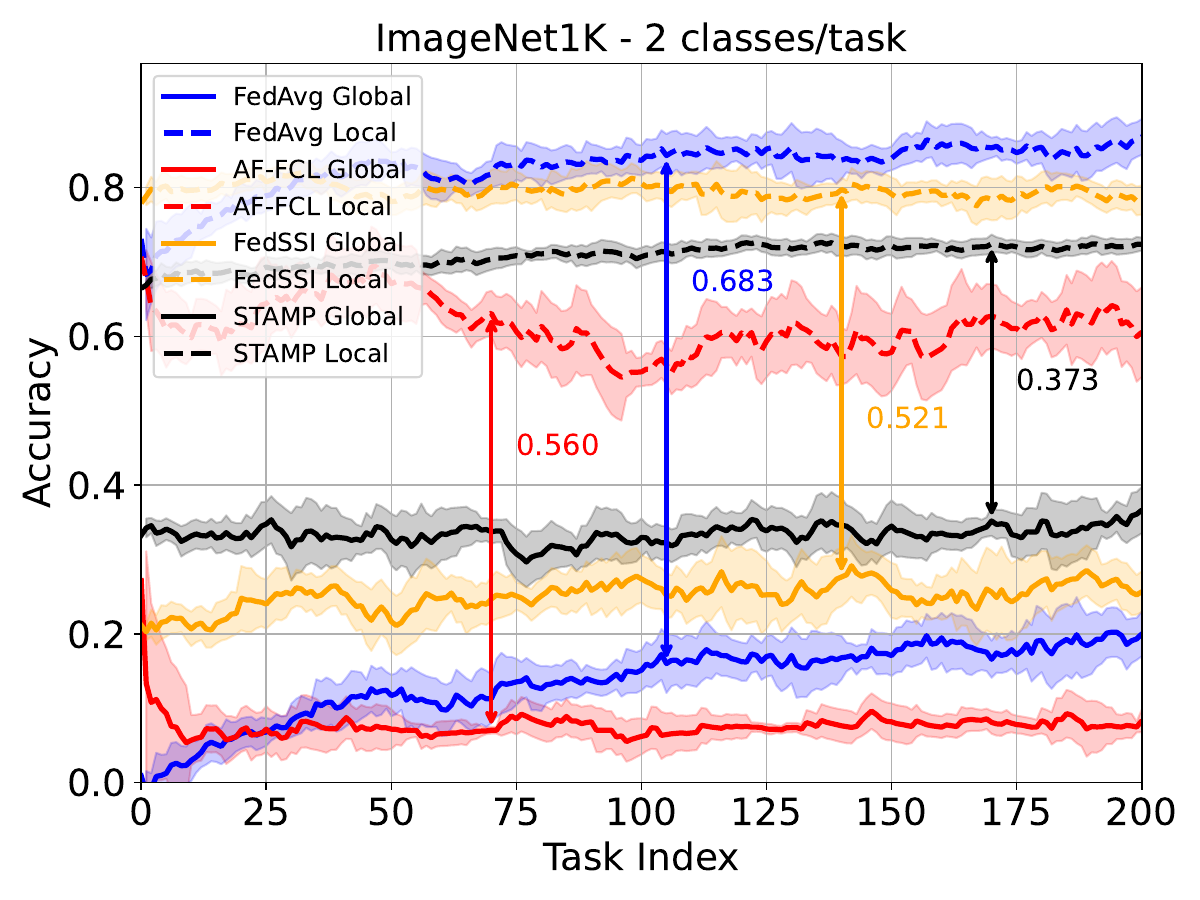}}
\caption{The illustration highlights the challenges encountered by current FCL methods (e.g., AF-FCL \citep{wuerkaixi2024accurate}, FedSSI \citep{2025-FCL-FedSSI}) when applied in heterogeneous settings. A notable gap between local and global test accuracy arises due to client-specific data heterogeneity within each task at every time step. STAMP demonstrates superior robustness over current baselines by mitigating inter-client divergence throughout the learning process, leading to a reduced local-global generalization gap.}
\label{fig:glr_radius}
\end{figure}
In FCL, clients collaboratively learn models for their private, sequential tasks while preserving data privacy. However, due to the sequential nature of these tasks, each client only has access to a limited amount of data from the current task. This constraint often leads to the loss of previously acquired knowledge, resulting in catastrophic forgetting.
Unlike existing studies, our focus is on a more challenging setting of FCL, coined heterogeneous FCL. This setting introduces two major challenges beyond those encountered in conventional FCL. First, in heterogeneous FCL, clients are engaged in non-identical tasks at any given time, resulting in a non-uniform learning environment (see Figure~\ref{fig:local-global}). As a consequence, the aggregated global model is subject to domain shifts at each communication round, impeding efficient convergence. Second, existing Federated Class-Incremental Learning (FCIL) methods perform poorly under heterogeneous FCL conditions. This is primarily because these methods typically assume a homogeneous class distribution across clients, enabling a straightforward model design where new task-specific heads are incrementally added. In contrast, heterogeneous FCL involves clients encountering different sets of classes at each task, leading to discrepancies in the architecture of local model heads. This variability substantially reduces the effectiveness of existing FCIL approaches in heterogeneous settings.

To address the aforementioned challenges, we propose a novel method, dubbed Federated Continual Learning via \textbf{\underline{S}}patio-\textbf{\underline{T}}emporal gr\textbf{\underline{A}}dient \textbf{\underline{M}}atching with network-free \textbf{\underline{P}}rototype (STAMP).
Our approach offers three key contributions:

First, we apply unified gradient matching across both spatial and temporal dimensions of the FCL system. Specifically, the spatial dimension refers to task differences across clients at a given time, while the temporal dimension refers to sequential tasks within a single client. By aligning gradients across both dimensions, our method enables the identification of aggregated gradients that minimize negative transfer both across sequential tasks and between clients, thereby improving knowledge retention and overall model performance.

Second, we utilize a prototypical network to generate prototypes that serve as stable gradient approximators, even in the presence of data perturbations. This design ensures more consistent gradient estimates of previous tasks when viewed through the lens of the current model—a property often overlooked in existing gradient-based approaches \citep{Luo_2023_CVPR, NIPS2017_f8752278, saha2021gradient, deng2021flattening}.

Third, we introduce a prototypical coreset selection strategy for efficient and robust prototype approximation. This approach offers two key advantages. (1) By carefully selecting a compact set of representative samples (coresets), our method maintains prototype quality and diversity over time with significantly reduced dependence on the prototypical networks or generative replay mechanisms used in prior work \citep{10378557, li2024an, chen2023saving, goswami2023fecam, 2023-FCL-FedCIL, NEURIPS2020_67d16d00}, both of which are prone to catastrophic forgetting in FCL settings. (2) The coreset-based strategy substantially reduces memory usage, as only a small number of representative samples per class per task are required to approximate the prototypes effectively.

\section{Backgrounds \& Preliminaries}
\subsection{Federated Continual Learning}
FCL refers to a practical learning scenario that melds the principles of FL and CL. Suppose that there are $U$ clients. 
On each client $u$, the model is trained on a sequence of $T$ tasks.
At a given step $t\times R + r$, where $R$ represents the number of communication rounds per task and $r$ is the current round of task $t$, client $u$ holds model parameters $\theta^{t,r}_{u}$ and only has access to the data from task $t$.
On client $u$, data $\gD^t_u$ of task $t$ consists of $N^t_u$ pairs of samples and their labels, i.e., $\gD^t_u = \{(x^t_i, y^t_i)^{N^t_u}_{i=1}\}$. 

In existing literature, the primary focus is on a specific task reshuffling setting, wherein the task set is identical for all clients, yet the arrival sequence of tasks differs \citep{2021-FCL-FedWeIT}. In practical scenarios, it may be observed that the task set of clients is not necessarily correlated. There is no guaranteed relation among the tasks $\{\gD^1_u, \gD^2_u, \ldots, \gD^T_u\}$ of client $u$ at different steps. Similarly, there is no guaranteed relation among the tasks $\{\gD^t_1, \gD^t_2, \ldots, \gD^t_U\}$ across different clients. Thus, we consider a more practical setting, the Limitless Task Pool (LTP). 

\textbf{Limitless Task Pool.} In the setting of LTP, tasks are selected randomly from a substantial repository of tasks, creating a situation where two clients may not share any common tasks (i.e., $ \{ \gD^i_u \}^{t_u}_{i=1} \cap \gD^i_v \}^{t_v}_{i=1} = \varnothing,~\forall u,v \in \{1,2,\ldots,U\}$). More importantly, clients possess diverse joint distributions of data and labels $p(x,y)$ due to statistical heterogeneity. Therefore, features learned from other clients could invariably introduce bias when applied to the current task of a client.

At every task $t$, our goal is to facilitate the collaborative construction of the global model with parameter $\theta^t$. Under the privacy constraint inherent in FL and CL, we aim to harmoniously learn current tasks while preserving performance on previous tasks for all clients, thereby seeking to optimize performance across all tasks seen so far by all clients as follows:
\begin{align}
    \min_{\theta^t}~[\gS^t_1, \gS^t_2, \ldots, \gS^t_U], \quad \textrm{where} \quad S^t_u = [\gL(\theta^t; \gD^1_u), \gL(\theta^t; \gD^2_u), \ldots, \gL(\theta^t; \gD^t_u)].
\end{align}
However, due to the resource limitation of distributed devices, the replay memory on clients are limited. Therefore, each client $u$, while performing the task $t$, does not have access to the samples of the previously learned task $\gD^{[1:t-1]}_u$. Therefore, the client model $\theta^t_u$ cannot be directly optimized to minimize the corresponding empirical risk $\sum^{t}_{i=1}\gL(\theta^t_u; \gD^i_u)$.
Moreover, data heterogeneity on each client at specific task $t$ introduces domain or label shifts, leading to discrepancies in data distributions across tasks and clients. This heterogeneity causes gradient conflict during training \citep{2025-FDG-FedOMG}.

\subsection{Gradient Matching}
When learning with various non-identical tasks, gradient conflict is one of the most critical issues.
\begin{definition}[Gradient conflict]
    The gradient $g_i$ and $g_j$ $(i\neq j)$ between two tasks $i, j$  are considered to be in conflict if their cosine similarity is negative, i.e., $\cos (g_i, g_j) = \frac{g_i \cdot g_j}{\vert g_i \vert \cdot \vert g_j \vert} < 0$. In this scenario, progress along the gradient $g_i$ results in negative transfer with respect to $g_j$, and vice versa.
\label{def:gradient-conflict}
\end{definition}
To mitigate the gradient conflict among tasks as in Definition~\ref{def:gradient-conflict}, we leverage the Gradient Matching (GM) approach proposed in \citep{2025-FDG-FedOMG} to achieve this objective
\begin{align}
    &\textrm{GM}(\rvg^{(r)}) = \frac{\kappa\Vert \Bar{g}^{(r)}\Vert}{\Vert \Gamma^*\rvg^{(r)}\Vert}\Gamma^*\rvg^{(r)} \notag \\
    \textrm{s.t.} \quad 
    &\Gamma^* = \arg\min_{\Gamma} \Gamma\rvg^{(r)}\cdot \Bar{g}^{(r)} + \kappa\Vert \Bar{g}^{(r)}\Vert\Vert g^{(r)}_{\Gamma}\Vert,
    \quad
    \Bar{g}^{(r)} = \frac{1}{\vert \gT \vert} \sum_{t\in\gT} g^{(r)}_t,
\end{align}
where $\rvg^{(r)} = [g^{(r)}_{t} \vert~t\in\gT]$ are the set of task-wise gradients, which participated in the training. The learned gradient $g_G = \textrm{GM}(\rvg^{(r)})$ utilizes the gradients of multiple tasks $\rvg^{(r)} = [g^{(r)}_{t} \vert~t\in\gT]$ to preserve the invariant properties of individual task-specific gradients. Specifically, since $g_G$ satisfies the condition $g_G \cdot g_i \leq 0$, $\forall i\in T$, it ensures that the resulting gradient does not induce negative transfer across tasks. Consequently, the aggregated gradient facilitates generalization across all tasks within the CL framework.

\section{Proposed Method}
\begin{figure}[htbp]
\vspace{-1mm}
    \centering
    \includegraphics[width=1\linewidth]{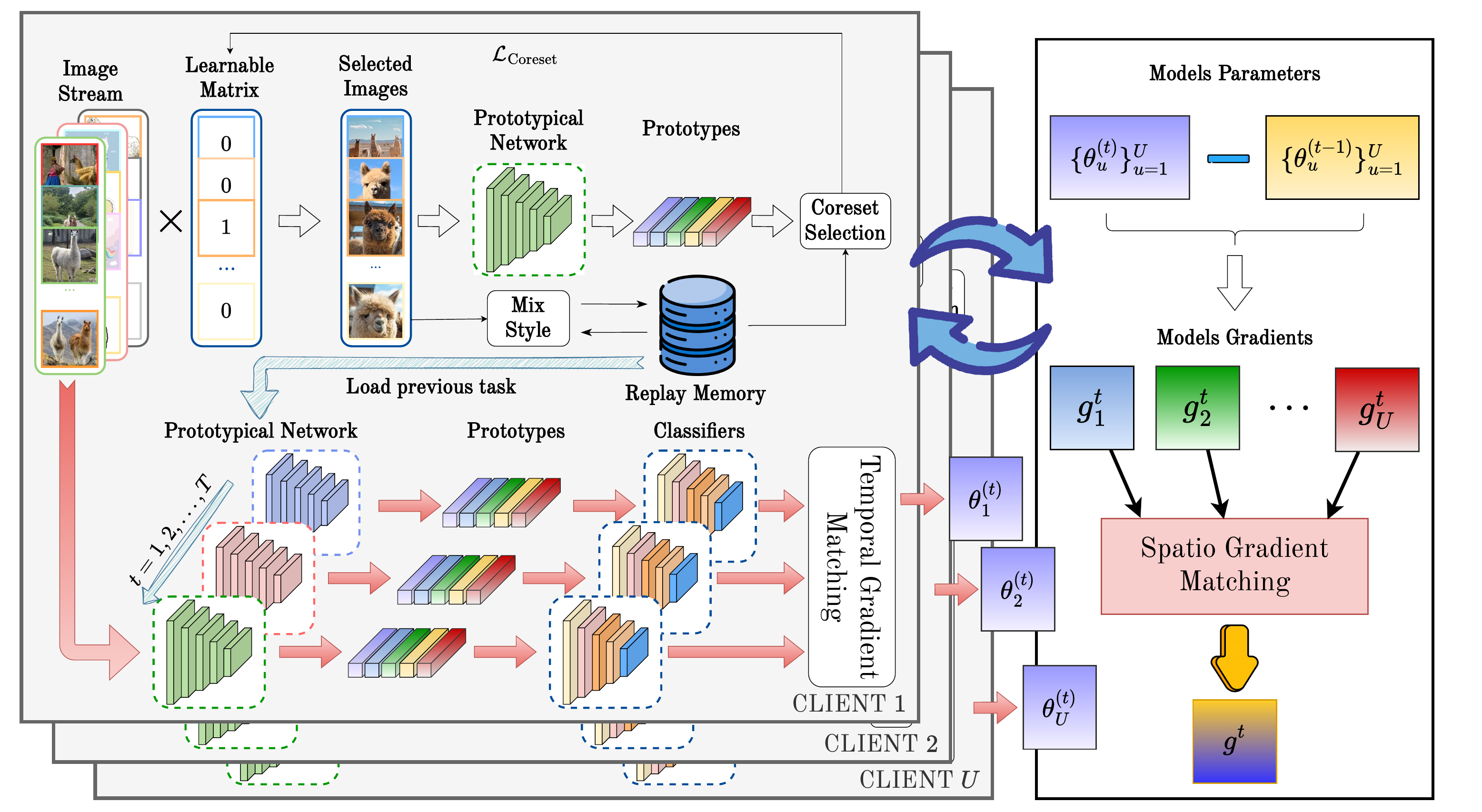}  
    \caption{Illustration of STAMP architecture.}
    \label{fig:STAMP-arch}
\vspace{-3mm}
\end{figure}

We introduce a novel framework, STAMP, for heterogeneous FCL. The STAMP framework (as shown in Figure~\ref{fig:STAMP-arch}) comprises three key components: (1) on-client temporal gradient matching, (2) on-server spatial gradient matching, (3) replay memory with network-free prototypes. 
\subsection{Temporal Gradient Matching}\label{sec:temporal-gm}
The temporal gradient matching technique is implemented on the client side in the local training. In particular, we take the gradients of previous tasks as input data for the gradient matching optimization problem as follows: 
\begin{align}
    \theta^{t, r+1}_u = \theta^{t, r}_u - \textrm{GM}(\rvg^{[0:t]}_u),
\end{align}
where $\rvg^{[0:t]}_u = [g^{i}_{u} \vert~i=\{1,2,\ldots,t\}]$ denotes the set of task-specific gradients, including the gradients of previous tasks $\rvg^{[0:t-1]}_u$ and current task $g^{t}_u$. Traditionally, the gradients from previous tasks are computed using stored data samples from past tasks to approximate the true gradients \citep{NIPS2017_f8752278, Luo_2023_CVPR, wu2024meta}. However, this approach requires a substantial memory buffer to store a sufficient amount of data for accurate gradient estimation. In scenarios where storage capacity is limited, the precision of the gradient approximation may be significantly compromised. An alternative solution to compute gradients is via prototype as follows: 
\begin{align}
    g^{(t)}_{u} = \frac{1}{C}\sum^C_{c=1}\nabla_{\theta^{t, r, E}_u} \gL\Big(f(p^{t}_{u,c}; \theta^{t, r, E}_u); c\Big).
\end{align}
To efficiently compute the prototypes for the gradient estimation, we employ the prototypical network \citep{NIPS2017_cb8da676}. However, the prototypical network and its continual counterpart \citep{10378557} may still suffer from catastrophic forgetting when deployed in the CL system.  
To mitigate this challenge, our intuition is to design prototypes that are learned without relying on prototype networks. To do so, we leverage a prototypical coreset which stores meaningful features for the prototype measurements in CL. The details of the prototypical coreset and its selection method are demonstrated in Section~\ref{sec:proto-coreset}.

\subsection{Spatio Gradient Matching}\label{sec:spatio-gm}
Building upon the work of \citep{2025-FDG-FedOMG}, the spatial gradient is computed on the server to identify a consistent gradient direction that remains invariant across heterogeneous tasks in FCL. This facilitates the global model in establishing a stable gradient direction, thereby mitigating the negative transfer that can occur due to task diversity. The update is given as follows:
\begin{align}
    \theta^{t, r+1} = \theta^{t, r} - \textrm{GM}(\rvg^{t}), \quad \rvg^{t} = [g^{t}_{u} \vert~u=\{1,2,\ldots,U\}],
\end{align}
where $\rvg^{t}$ represents the collection of local gradients obtained from the participating clients. Each local gradient is computed as $g^{t}_{u} = \theta^{t,r+1}_u - \theta^{t,r}_u$, using the model updates, and thus incurs no additional communication overhead. By aligning the gradient directions across clients, this method effectively addresses task heterogeneity, reducing the detrimental impact of client drift in heterogeneous FCL.

\subsection{Prototypical Coreset assisted Replay Memory}\label{sec:proto-coreset}
\paragraph{Prototypical Coreset Selection}
Our primary objective is to identify salient samples within $\gM^l$ for each label $l$ such that their combined representations, as processed by the encoder $\phi$ form a prototype on class $l$. To achieve this, we aim to select $\vert\gM^l\vert$ samples from the current tasks corresponding to label $l$. The selected samples are then utilized to transfer information to network-free prototypes through MixStyle \citep{zhou2021domain}. The selection process is defined as follows:
\begin{align}
    \widetilde{X}^l &= \argmin_{A} \Big\Vert \Big[\frac{1}{\vert\gM^l\vert}\sum^{}_{i\in\gM^l} g(x_i; \phi) + \frac{1}{\vert\gN^{t}_l\vert}\sum^{}_{i\in\gN^{t}_l} a_i \cdot g(x_i; \phi)\Big] - p^l \Big\Vert^2, \\
    \textrm{ s.t. }~~ p^l &= \frac{1}{\sum^T_{t=1} \vert\gN^t_l\vert} \Big[g(\widetilde{x}^l; \phi)\cdot\sum^{t-1}_{j=1} \vert\gN^j_l\vert + \sum^{}_{i\in\gN^{t}_l} g(x_i; \phi)\Big] \cdot \mathbbm{1}\{y_j = l\},  \notag \\ 
    X^l &= \{ x_i \mid a_i \in A \}, ~~ X^l = K. \notag 
\label{eq:network-free-proto}
\end{align}
Here, $\vert\gM^l\vert$ is a number of samples can be stored in a replay memory according to class $l$ and a pre-defined hyperparameter in the FCL system. Our objective is to select a set of $K$ samples from both the network-free prototype learned from previous tasks and the newly acquired data from the current task $t$. If the number of selected samples exceeds $K$, we apply MixStyle to blend the style of the newly selected data with that of the previously identified samples, as formulated below:
\begin{align}
    \textrm{MixStyle}(\widetilde{x}^l; x) &= \gamma_\textrm{mix} \frac{\widetilde{x}^l - \mu(\widetilde{x}^l)}{\sigma(\widetilde{x}^l)}+ \beta_\textrm{mix}, \\
    \textrm{ s.t. }~~ \gamma_\textrm{mix} &= \lambda\sigma(\widetilde{x}^l) + (1-\lambda)\sigma(x)
                  ,~~\beta_\textrm{mix}   = \lambda\mu(\widetilde{x}^l) + (1-\lambda)\mu(x),    \notag
\end{align}
where $x$ are the newly satisfying network-free prototype found from \eqref{eq:network-free-proto}. To make the encoder $\phi$ learn the prototype better, we inherit the prototypical network \citep{NIPS2017_cb8da676} learning process to learn the encoder $\phi$.

\begin{table}[h!]
\centering
\caption{We report the average per-task performance of FCL under a setting where each task is assigned 20 classes. Evaluations are conducted using 10 clients (fraction = $1.0$) across 5 independent trials. OOM refers to the out of memory in GPU. $\uparrow$ and $\downarrow$ indicate that higher and lower values are better, respectively. C$\rightarrow$S and S$\rightarrow$C denote communication from the client to the server and from the server to the client, respectively.}
\label{tab:baselines-task20}
\footnotesize
\setlength\tabcolsep{3pt}
\begin{tabular}{lccccccc}
\toprule
\multicolumn{8}{c}{\textbf{CIFAR100} ($U=10$, $C=20$)} \\
\midrule
\textbf{Methods} & \textbf{Accuracy} $\uparrow$ & \textbf{AF} $\downarrow$ & \textbf{Avg. Comp.} $\downarrow$ & \multicolumn{2}{c}{\textbf{Comm. Cost} $\downarrow$}  & \textbf{GPU (Peak)} $\downarrow$ & \textbf{Disk} $\downarrow$ \\
  &   &   & \textbf{(Sec/Round)} & \textbf{C$\rightarrow$S} & \textbf{S $\rightarrow$ C} &  & \\
\midrule
FedAvg & \pmval{27.2}{2.2} & \pmval{5.9}{0.9} & 27.6 sec & 44.6 MB & 44.6 MB & 1.92 GB & N/A \\
FedALA & \pmval{28.5}{2.4} & \pmval{6.5}{1.2} & 28.2 sec & 44.6 MB & 44.6 MB & 1.93 GB & N/A \\
FedDBE & \pmval{28.3}{1.6} & \pmval{5.5}{0.7} & 28.3 sec & 44.6 MB & 44.6 MB & 1.91 GB & N/A \\
FedAS & \pmval{40.2}{1.1} & \pmval{30.7}{0.3} & 135.7 sec & 44.6 MB & 44.6 MB & 1.92 GB & N/A \\
FedOMG & \pmval{36.8}{1.4} & \pmval{8.5}{0.6} & 32.7 sec & 44.6 MB & 44.6 MB & 1.92 GB & N/A \\
\midrule
GLFC & \pmval{29.8}{2.1} & \pmval{7.5}{0.4} & 167.8 sec & 88.2 MB & 46.5 MB & 3.83 GB & 22.1 MB \\
FedCIL & \pmval{32.4}{1.7} & \pmval{6.3}{1.2} & 199.3 sec & 95.3 MB & 44.6 MB & 4.21 GB & 18.5 MB \\
LANDER & \pmval{45.1}{1.3} & \pmval{5.4}{0.8} & 153.6 sec & 112.4 MB & 138.7 MB & 4.83 GB & 131.5 MB \\
TARGET & \pmval{32.1}{2.3} & \pmval{5.9}{1.6} & 236.4 sec & 112.4 MB & 44.6 MB & 3.65 GB & 18.5 MB \\
FedL2P & \pmval{30.2}{1.8} & \pmval{6.3}{1.3} & 78.1 sec & 56.3 MB & 56.3 MB & 2.56 GB & N/A \\
\midrule
FedWeIT & \pmval{37.3}{2.3} & \pmval{4.7}{0.8} & 38.7 sec & 44.2 MB & 44.2 MB & 7.21 GB & N/A \\
AF-FCL & \pmval{35.6}{0.4} & \pmval{5.2}{0.5} & 45.3 sec & 156.3 MB & 121.3 MB & 8.93 GB & N/A \\
\midrule
\textbf{STAMP}      & \pmval{41.3}{0.9} & \pmval{5.4}{0.6} & 56.3 sec & 44.6 MB & 44.6 MB & 1.92 GB & 16.3 MB \\
\midrule
\midrule
\multicolumn{8}{c}{\textbf{ImageNet1K} ($U=10$, $C=20$)} \\
\midrule
FedAvg & \pmval{17.3}{3.3} & \pmval{14.1}{0.2} & 1485.2 sec & 112.5 MB & 112.5 MB & 16.11 GB & N/A \\
FedALA & \pmval{17.6}{5.6} & \pmval{14.9}{0.8} & 1556.6 sec & 112.5 MB & 112.5 MB & 16.12 GB & N/A \\
FedDBE & \pmval{18.8}{5.2} & \pmval{13.9}{0.3} & 1572.7 sec & 112.5 MB & 112.5 MB & 16.11 GB & N/A \\
FedAS & \pmval{22.3}{5.0} & \pmval{18.2}{0.6} & 5108.5 sec & 112.5 MB & 112.5 MB & 16.11 GB & N/A \\
FedOMG & \pmval{21.2}{3.3} & \pmval{11.3}{0.7} & 1821.2 sec & 112.5 MB & 112.5 MB & 16.11 GB & N/A \\
\midrule
GLFC & \pmval{22.5}{2.1} & \pmval{6.3}{0.2} & 5647.3 sec & 225.3 MB & 121.2 MB & 20.24 GB & 112.6 MB \\
FedCIL & \pmval{24.1}{2.8} & \pmval{7.3}{0.4} & 7120.3 sec & 245.5 MB & 112.5 MB & 23.47 GB & 184.3 MB \\
LANDER & \pmval{31.8}{1.4} & \pmval{7.8}{0.9} & 6825.8 sec & 267.4 MB & 453.6 MB & 26.54 GB & 1.31 GB \\
TARGET & \pmval{25.8}{3.8} & \pmval{6.7}{0.4} & 9958.2 sec & 287.4 MB & 112.5 MB & 21.08 GB & 184.3 MB \\
FedL2P & \pmval{22.3}{3.7} & \pmval{9.4}{0.6} & 3278.7 sec & 146.6 MB & 146.6 MB & 18.21 GB & N/A \\
\midrule
FedWeIT & \pmval{24.8}{1.3} & \pmval{5.1}{0.8} & 1763.8 sec & 110.4 MB & 110.4 MB & 41.23 GB & 61.7 GB \\
AF-FCL & \pmval{21.3}{5.1} & \pmval{4.5}{0.6} & 1823.7 sec & 421.3 MB & 336.8 MB & 46.81 GB & N/A \\
\midrule
\textbf{STAMP} & \pmval{26.8}{2.3} & \pmval{5.8}{0.4} & 3041.2 sec & 112.5 MB & 112.5 MB & 16.11 GB & 152.6 MB \\
\bottomrule
\end{tabular}
\vspace{-3mm}
\end{table}

\paragraph{Prototypical Network with Coreset}
On each client $u$, the prototype $p^{t}_{u,l}$ on label $l$ are computed via a prototypical network \citep{NIPS2017_cb8da676} via 
$p^{t}_{u,l} = \frac{1}{\vert \gD^t_{u,l}\vert}\sum_{x_i\in\gD^t_{u,l}} g(x_i; \phi)$. The prototypical network is learned via a loss function as follows:
\begin{align}
    \phi^* = \argmin_\phi \sum^{L}_{l=1} d\Big(g(x; \phi), p_l\Big) + \log\sum_{l'}\exp\Big(d\Big(g(x; \phi), p_l\Big)\Big).
\label{eq:protoloss}
\end{align}
The objective of \eqref{eq:protoloss} is to ensure that the learned prototype $g(x;\phi)$, derived from the input data $x$, closely aligns with the computed prototype of the same class $l$ across the entire batch, while simultaneously maintaining a significant distance from approximated prototypes of other classes $l'$.
\section{Experimental Evaluations}\label{sec:experiments}
In this section, we conduct extensive experiments to demonstrate the effectiveness of STAMP. The implementation details and additional experiments are provided in Appendices~\ref{app:settings}
. To ensure a fair assessment of FCL baselines under heterogeneous settings and catastrophic forgetting, we do not use pretrained models, as their training data (e.g., ImageNet1K) overlaps with our dataset, potentially biasing the evaluation.

\subsection{Benchmarking with Baselines}
\begin{table}[h!]
\centering
\caption{We report the average per-task performance of FCL under a setting where each task is assigned 2 classes. Evaluations are conducted using 10 clients (fraction = $1.0$) across 5 independent trials. OOM refers to the out of memory in GPU. $\uparrow$ and $\downarrow$ indicate that higher and lower values are better, respectively. C$\rightarrow$S and S$\rightarrow$C denote communication from the client to the server and from the server to the client, respectively.}
\label{tab:baselines-task2}
\footnotesize
\setlength\tabcolsep{3pt}
\begin{tabular}{lccccccc}
\toprule
\multicolumn{8}{c}{\textbf{CIFAR100} ($U=10$, $C=2$)} \\
\midrule
\textbf{Methods} & \textbf{Accuracy} $\uparrow$ & \textbf{AF} $\downarrow$ & \textbf{Avg. Comp.} $\downarrow$ & \multicolumn{2}{c}{\textbf{Comm. Cost} $\downarrow$}  & \textbf{GPU (Peak)} $\downarrow$ & \textbf{Disk} $\downarrow$ \\
  &   &   & \textbf{(Sec/Round)} & \textbf{C$\rightarrow$S} & \textbf{S $\rightarrow$ C} &  &  \\
\midrule
FedAvg & \pmval{31.7}{1.7} & \pmval{25.2}{1.3} & 3.3 sec & 44.6 MB & 44.6 MB & 1.92 GB & N/A \\
FedALA & \pmval{36.5}{2.4} & \pmval{27.3}{0.5} & 3.6 sec & 44.6 MB & 44.6 MB & 1.93 GB & N/A \\
FedDBE & \pmval{37.0}{1.6} & \pmval{26.1}{0.7} & 3.6 sec & 44.6 MB & 44.6 MB & 1.91 GB & N/A \\
FedAS & \pmval{58.2}{0.1} & \pmval{56.1}{0.1} & 13.7 sec & 44.6 MB & 44.6 MB & 1.92 GB & N/A \\
FedOMG & \pmval{39.1}{1.3} & \pmval{24.5}{0.4} & 4.1 sec & 44.6 MB & 44.6 MB & 1.92 GB & N/A \\
\midrule
GLFC & \pmval{44.8}{2.1} & \pmval{29.5}{0.4} & 18.3 sec & 88.2 MB & 46.5 MB & 4.33 GB & 22.1 MB \\
FedCIL & \pmval{46.5}{2.2} & \pmval{28.8}{1.2} & 22.3 sec & 95.3 MB & 44.6 MB & 4.81 GB & 18.5 MB \\
LANDER & \pmval{50.8}{1.3} & \pmval{22.6}{0.4} & 15.8 sec & 88.2 MB & 104.3 MB & 5.26 GB & 131.5 MB \\
TARGET & \pmval{45.1}{2.4} & \pmval{28.6}{1.6} & 25.6 sec & 112.4 MB & 44.6 MB & 3.65 GB & 18.5 MB \\
FedL2P & \pmval{48.2}{1.8} & \pmval{28.1}{0.6} & 8.6 sec & 56.3 MB & 56.3 MB & 2.56 GB & N/A \\
\midrule
FedWeIT & \pmval{52.6}{1.3} & \pmval{25.7}{0.9} & 5.4 sec & 44.5 MB & 44.5 MB & 5.83 GB & 61.7 GB \\
AF-FCL & \pmval{51.4}{0.7} & \pmval{48.7}{1.2} & 4.9 sec & 156.3 MB & 121.3 MB & 8.93 GB & N/A \\
\midrule
\textbf{STAMP} & \pmval{52.8}{0.9} & \pmval{24.3}{0.8} & 9.1 sec & 44.6 MB & 44.6 MB & 1.92 GB & 16.3 MB \\
\midrule
\midrule
\multicolumn{8}{c}{\textbf{ImageNet1K} ($U=10$, $C=2$)} \\
\midrule
FedAvg & \pmval{24.3}{5.1} & \pmval{19.6}{0.1} & 133.2 sec & 112.5 MB & 112.5 MB & 16.11 GB & N/A \\
FedALA & \pmval{27.2}{9.1} & \pmval{20.3}{0.2} & 141.6 sec & 112.5 MB & 112.5 MB & 16.12 GB & N/A \\
FedDBE & \pmval{29.2}{7.2} & \pmval{19.4}{0.2} & 142.7 sec & 112.5 MB & 112.5 MB & 16.11 GB & N/A \\
FedAS & \pmval{43.5}{4.4} & \pmval{40.2}{0.4} & 498.5 sec & 112.5 MB & 112.5 MB & 16.11 GB & N/A \\
FedOMG & \pmval{30.4}{3.8} & \pmval{21.1}{0.7} & 171.3 sec & 112.5 MB & 112.5 MB & 16.11 GB & N/A \\
\midrule
GLFC & \pmval{31.4}{3.1} & \pmval{27.4}{0.6} & 466.7 sec & 225.3 MB & 121.2 MB & 20.24 GB & 112.6 MB \\
FedCIL & \pmval{33.8}{3.6} & \pmval{25.8}{0.7} & 652.3 sec & 245.5 MB & 112.5 MB & 23.47 GB & 184.3 MB \\
LANDER & \pmval{34.9}{2.7} & \pmval{26.1}{0.9} & 573.8 sec & 573.8 sec & 453.6 MB & 26.54 GB & 1.31 GB \\
TARGET & \pmval{33.2}{4.2} & \pmval{25.2}{0.4} & 913.2 sec & 287.4 MB & 112.5 MB & 21.08 GB & 184.3 MB \\
FedL2P & \pmval{34.5}{4.8} & \pmval{26.4}{0.2} & 303.7 sec & 146.6 MB & 146.6 MB & 18.21 GB & N/A \\
\midrule
FedWeIT & \pmval{39.7}{3.1} & \pmval{21.5}{-} & 194.2 sec & 111.8 MB & 111.8 MB & 62.7 GB & 640 GB \\
AF-FCL & \pmval{8.3}{5.3} & \pmval{46.6}{0.3} & 176.7 sec & 421.3 MB & 336.8 MB & 46.81 GB & N/A \\
\midrule
\textbf{STAMP} & \pmval{41.5}{2.4} & \pmval{24.2}{0.8} & 321.2 sec & 112.5 MB & 112.5 MB & 16.11 GB & 152.6 MB \\
\bottomrule
\end{tabular}
\vspace{-2mm}
\end{table}

Tables~\ref{tab:baselines-task20} and \ref{tab:baselines-task2} report results on the CIFAR100 and ImageNet1K datasets, both featuring varying class distributions across tasks. In addition to average accuracy and average forgetting (AF), we assess key system-level metrics: computational overhead, communication cost, GPU utilization, and disk usage. Computational overhead is measured as the average time per round, reflecting the cost of client-side training, especially for generative models. Communication cost denotes the average data transferred (in GB) per client-server round. GPU utilization captures peak memory usage, critical in resource-limited settings, while disk usage reflects the total client-side storage required, including replay buffers and task-specific model parameters.
The vanilla FL baselines, e.g., FedAvg, FedALA, FedAS, FedDBE, and FedOMG, may lead the model easily to forget the knowledge from past tasks, as indicated by high average forgetting. 
\begin{figure}[htbp]
\vspace{-1mm}
    \centering
    \includegraphics[width=\linewidth]{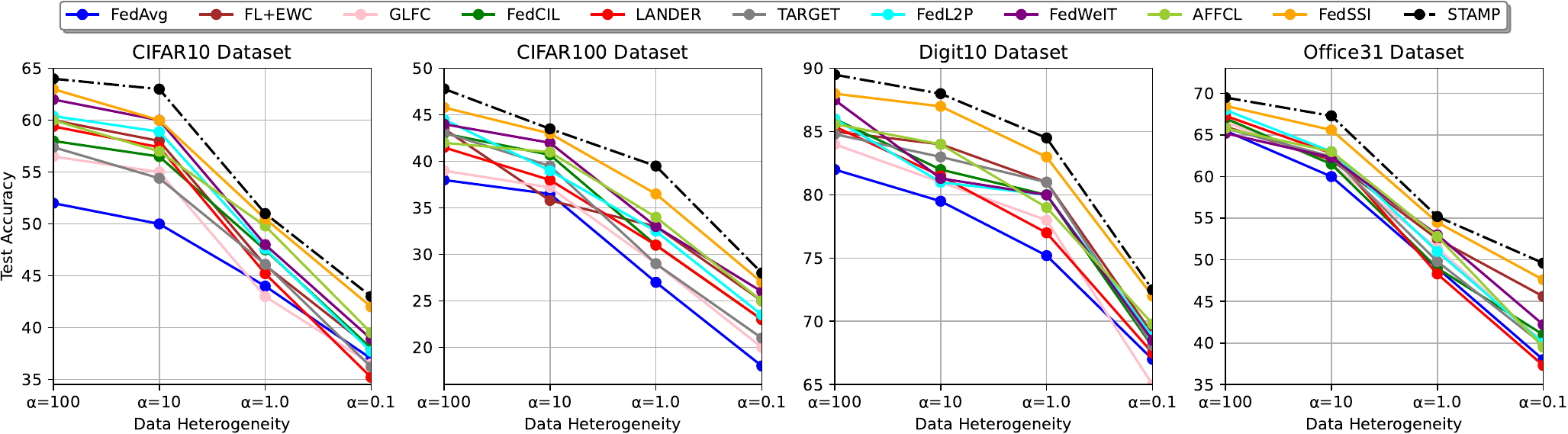}  
    \caption{Performance w.r.t data heterogeneity $\alpha$ for four datasets.}
    \label{fig:data-heterogeneity}
\vspace{-3mm}    
\end{figure}
FedWeIT\footnote[1]{ The official code of FedWeIT can be found at: \url{https://github.com/wyjeong/FedWeIT}.} stores task-specific head parameters in GPU memory. However, when both the number of classes (e.g., $1000$ classes in ImageNet1K) and the number of tasks (e.g., $500$ tasks in our ImageNet setup) become large, the total number of parameters grows significantly\footnote[2]{We observe from the official code that FedWeIT needs more than 512 GB of RAM memory to be able to run a simple LeNet on ImageNet. As such, we have to save the task-adaptive parameters in memory. In our reformatted implementation, we mitigate this memory constraint by utilizing disk storage for model loading.}. As a result, storing all task-specific parameters in GPU memory becomes infeasible, and they must instead be saved to disk. However, this approach leads to a substantial increase in average training time. 
LANDER stores all generated pseudo task-specific data on disk, incurring client-side storage overhead comparable to conventional CL methods using replay memory. Additionally, broadcasting synthetic data from the server to clients introduces substantial communication overhead.

The key observations from Tables~\ref{tab:baselines-task20} and \ref{tab:baselines-task2} indicate that more challenging settings—specifically, those with only two classes per task—exhibit greater susceptibility to catastrophic forgetting. This is because each task provides less comprehensive information about the overall dataset, thereby leading to a higher average forgetting (AF) score. STAMP demonstrates the most substantial improvements in two key metrics: accuracy and forgetting. Moreover, its communication cost remains comparable to that of standard FL. Additionally, STAMP requires relatively modest RAM and disk resources, making it suitable for deployment on resource-constrained devices.
\subsection{Performance under tasks with non-IID settings} Figure~\ref{fig:data-heterogeneity} illustrates the test accuracy across varying levels of data heterogeneity for CIFAR10, CIFAR100, Digit10, and Office31 datasets. As shown in the figure, all methods improve test accuracy as data heterogeneity decreases (i.e., larger $\alpha$). Notably, STAMP consistently achieves superior and stable performance across different levels of heterogeneity, indicating its robustness under non-IID conditions.

\subsection{Analysis on STAMP}
\subsubsection{Efficiency of Prototypical Coreset}
\begin{figure}[htbp]
    \centering
    \includegraphics[width=\linewidth]{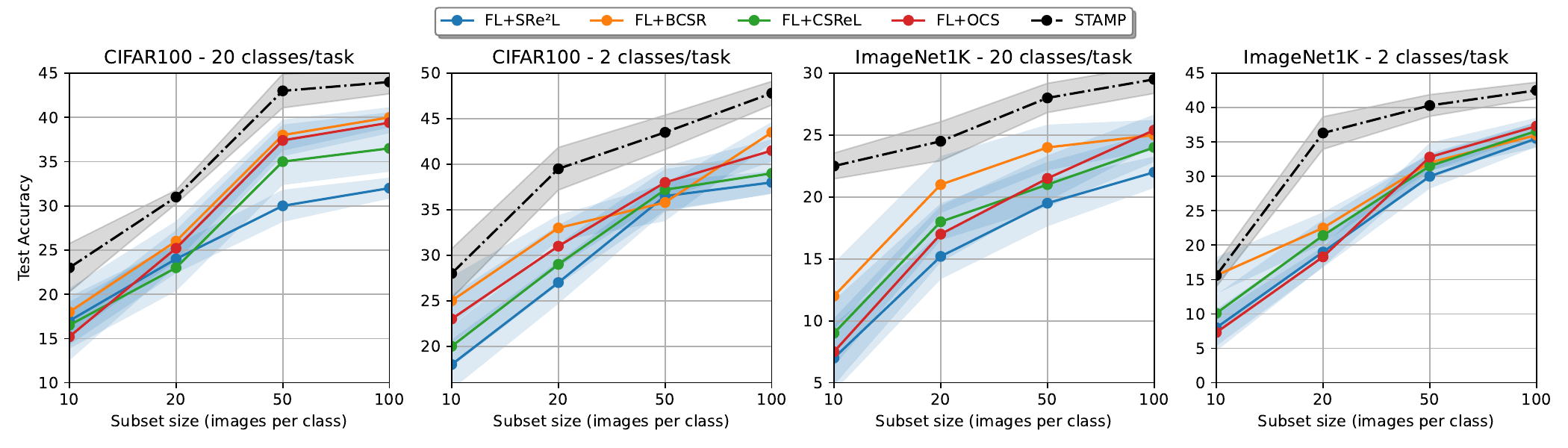}  
    \caption{Performance comparisons in coreset selection demonstrate that our approach outperforms the integration of alternative baseline methods within vanilla FL.}
    \label{fig:protocore}
\vspace{-5mm}    
\end{figure}

To evaluate the effectiveness of our proposed coreset selection method, we compare STAMP with a vanilla FL framework incorporating alternative data condensation techniques on the client side, including SRe$^2$L \citep{yin2023squeeze}, BCSR \citep{hao2023bilevel}, and OCS \citep{yoon2022online}, CSReL \citep{tong2025coreset}. The experimental results in Figure~\ref{fig:protocore} show that our method consistently outperforms these coreset selection-based FL algorithms. Notably, our approach can reduce the coreset size to as few as $20$ images per class without significantly compromising performance compared to training on the full-scale dataset for previous tasks.

\subsubsection{Efficiency of Temporal Gradient Matching}
\begin{figure}[ht]
\vspace{-7mm}
\centering
\subfloat[\label{fig:CIFAR100-20C}]{\includegraphics[width=0.28\linewidth]{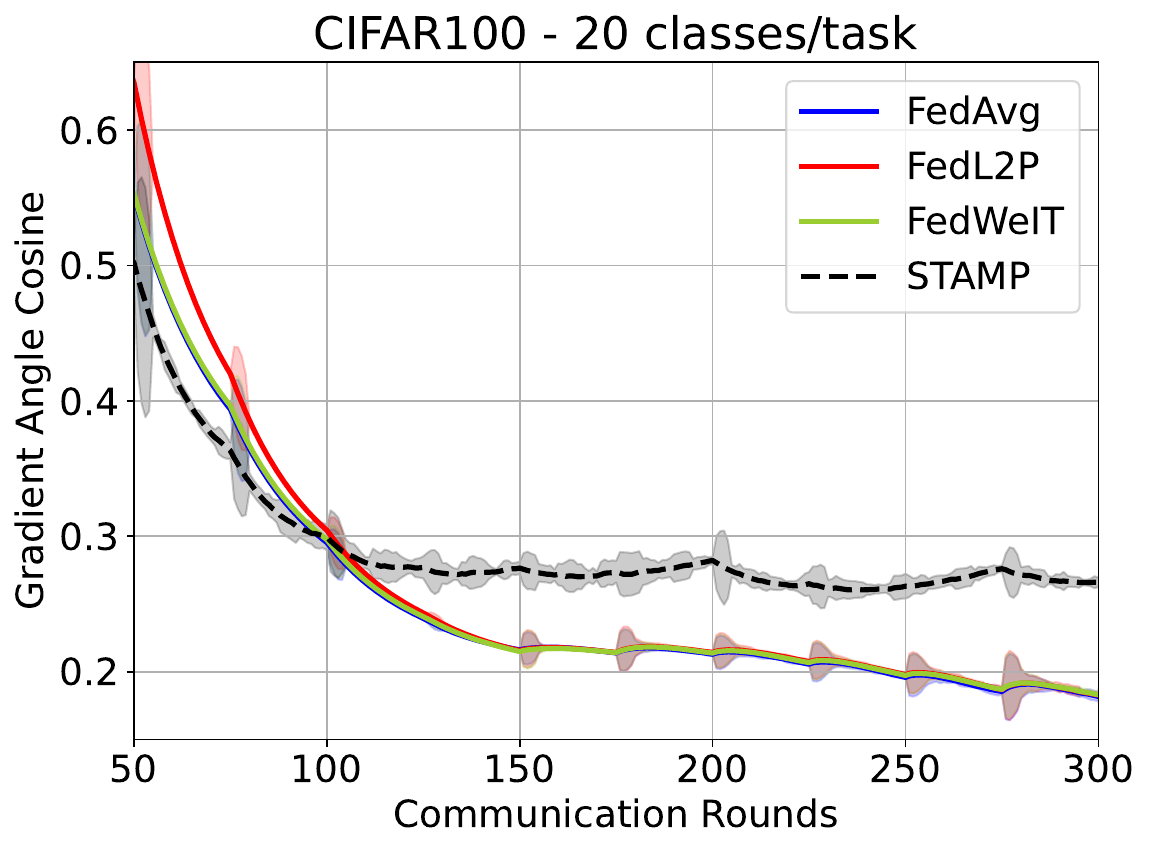}}
\subfloat[\label{fig:ImageNet1K-2C}]{\includegraphics[width=0.28\linewidth]{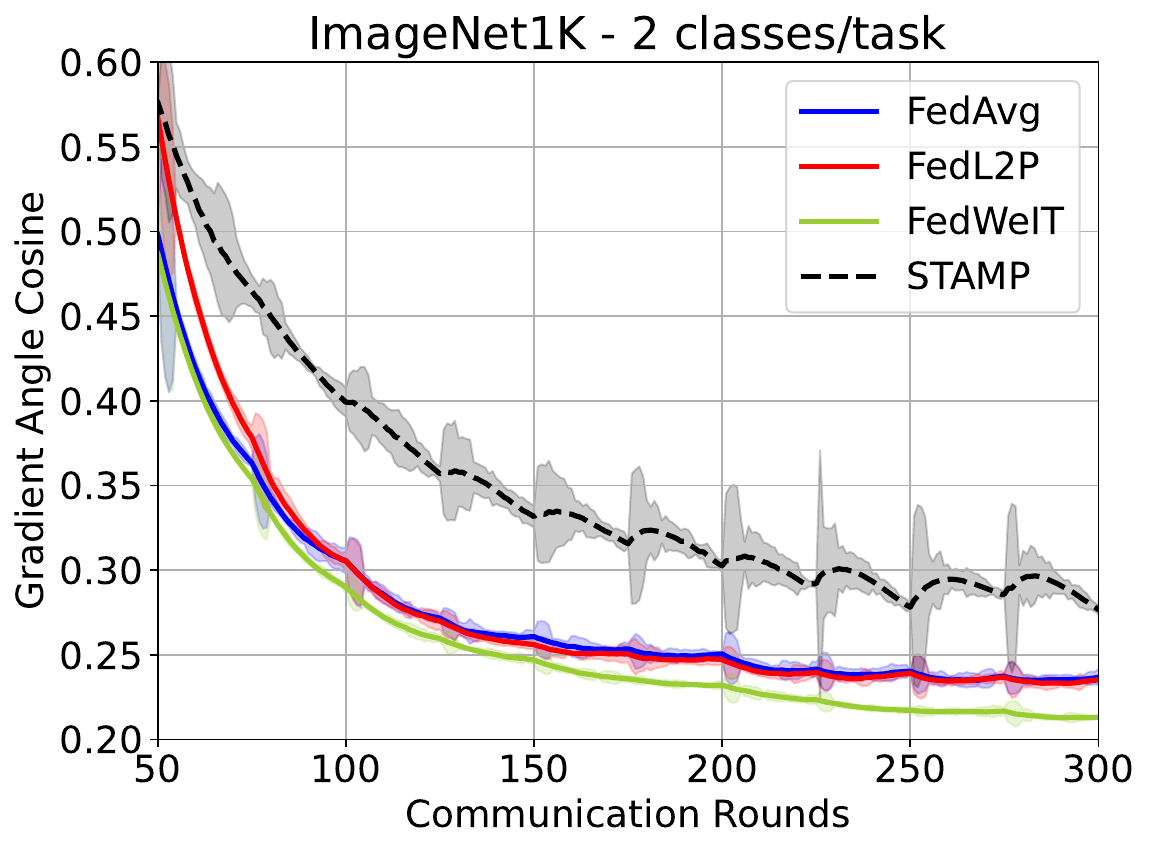}}\;
\subfloat[\label{fig:ImageNet1K-20C}]{\includegraphics[width=0.28\linewidth]{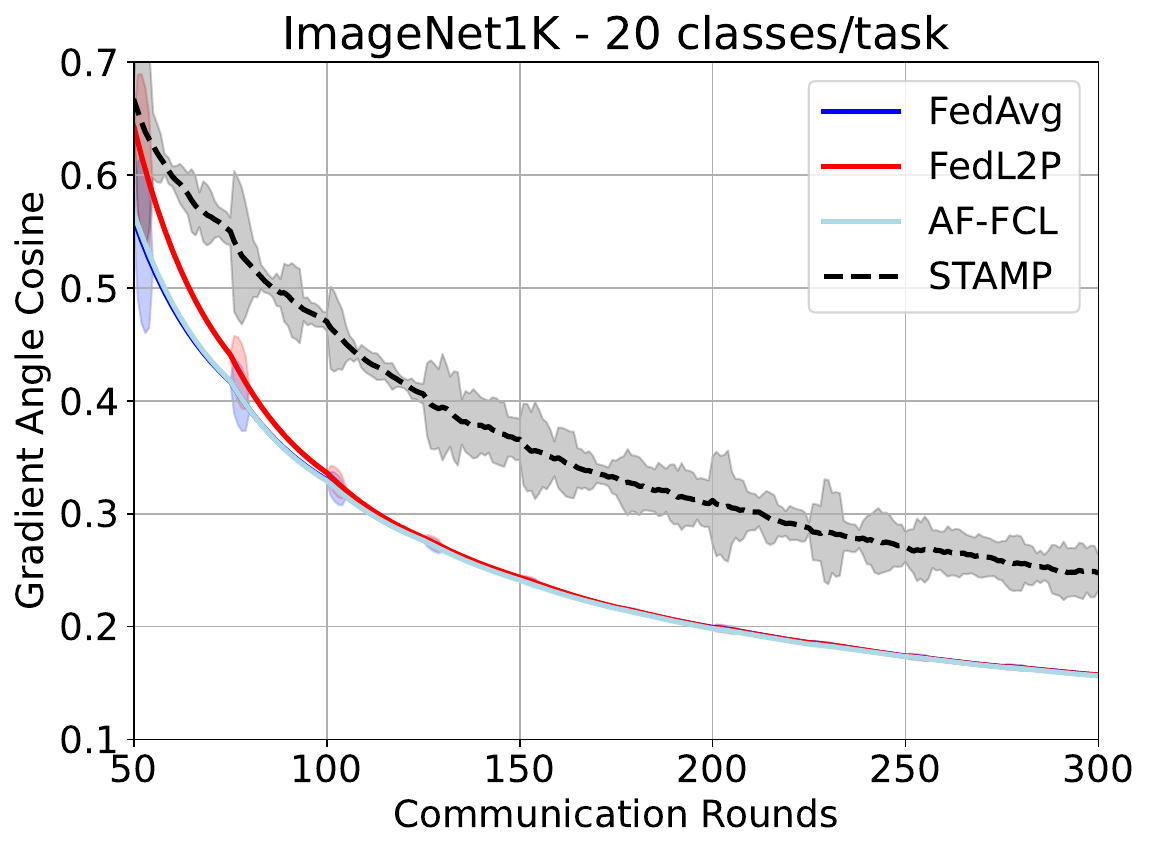}}
\caption{The figures illustrate the average temporal gradient angles across different baseline methods. Specifically, Figure~\ref{fig:CIFAR100-20C} shows the gradient cosine similarity on CIFAR100 under a 20 classes per task setting. Figure~\ref{fig:ImageNet1K-2C} presents the gradient cosine similarity for ImageNet1K with 2 classes per task, and Figure~\ref{fig:ImageNet1K-20C} depicts the results for ImageNet1K under a 20 classes per task configuration.}
\vspace{-5mm}
\label{fig:temporal-angle}
\end{figure}

To evaluate the effectiveness of temporal gradient matching on the client side, we analyze the gradient angles produced by STAMP on CIFAR100 and ImageNet1K datasets and compare them with two sets of baseline methods: FedAvg and FedL2P for standard FL, and FedWeIT and AF-FCL, for FCL. The results are presented in Figure~\ref{fig:temporal-angle}. As shown, STAMP demonstrates superior gradient alignment with previously learned tasks. This improved alignment suggests that STAMP is less prone to catastrophic forgetting compared to existing approaches.

\subsubsection{Efficiency of Spatio Gradient Matching}
\begin{figure}[ht]
\centering
\subfloat[\label{fig:cosine-cifar100-10c}]{\includegraphics[width=0.28\linewidth]{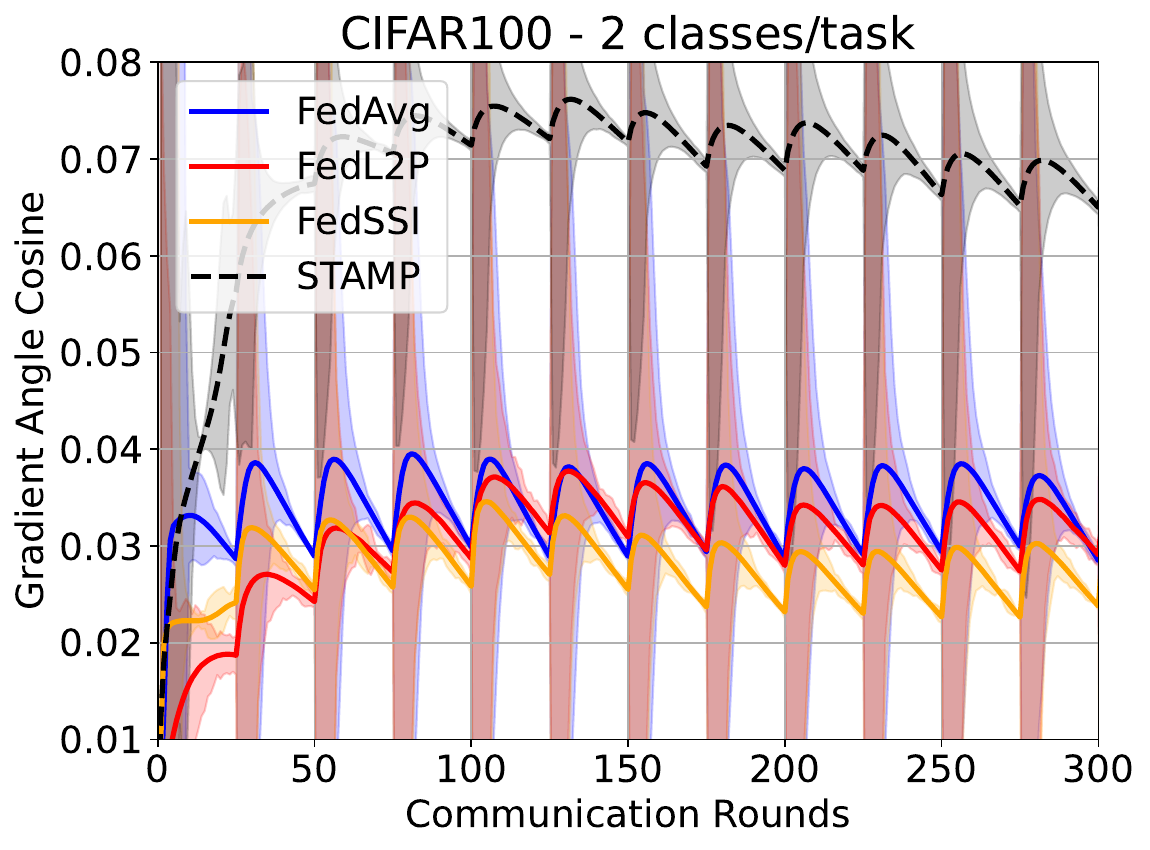}}\;
\subfloat[\label{fig:cosine-imagenet1k-20c}]{\includegraphics[width=0.28\linewidth]{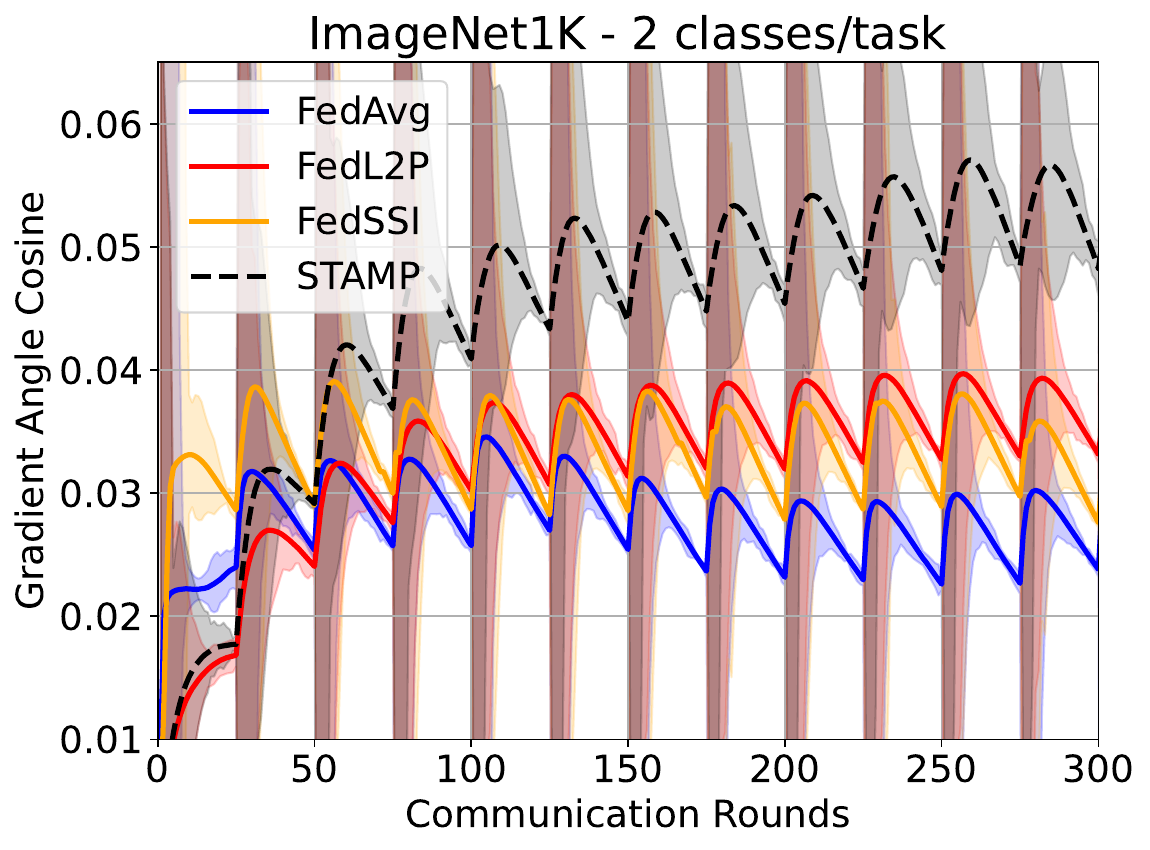}}\;
\subfloat[\label{fig:cosine-imagenet1k-2c}]{\includegraphics[width=0.28\linewidth]{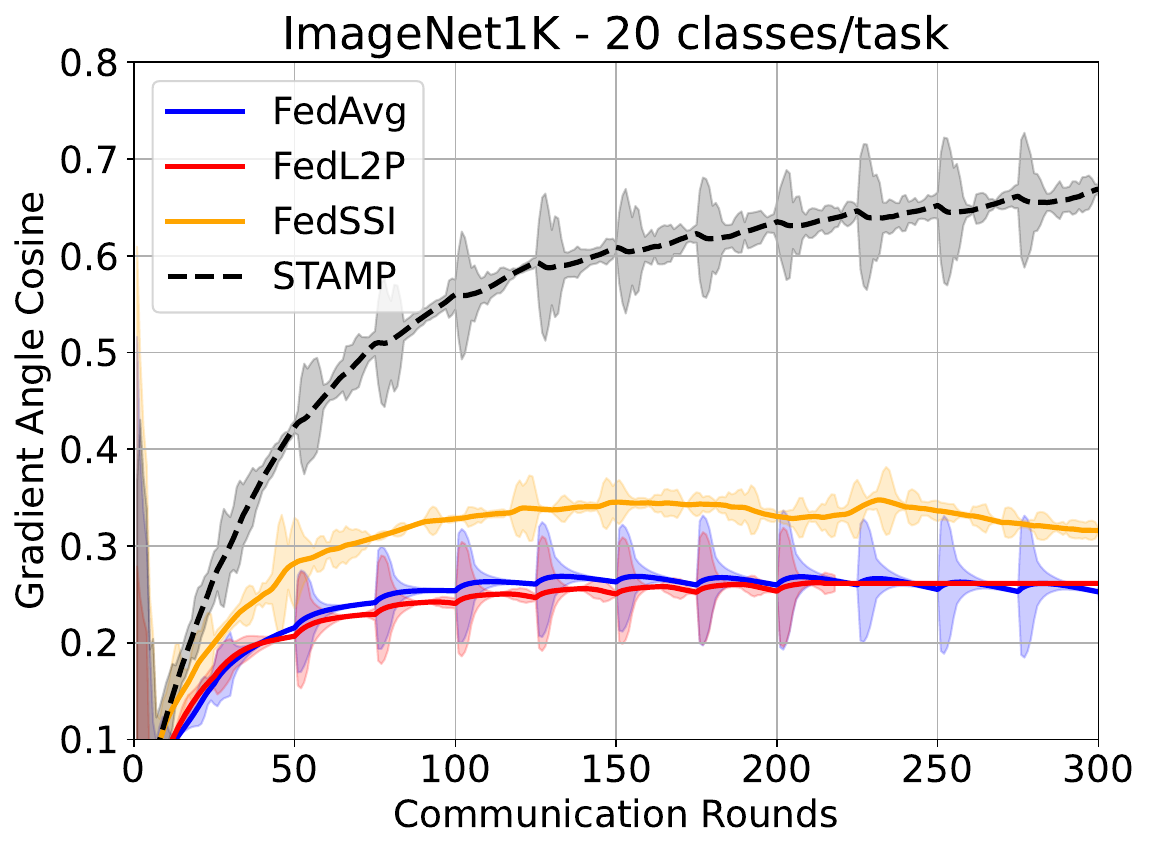}}
\caption{The figures illustrate the average spatio gradient angles across different baseline methods. Specifically, Figure~\ref{fig:cosine-cifar100-10c} shows the gradient cosine similarity on CIFAR100 under a 2 classes per task setting. Similarly, Figure~\ref{fig:cosine-imagenet1k-20c} presents the gradient cosine similarity for ImageNet1K with 2 classes per task, and Figure~\ref{fig:cosine-imagenet1k-2c} depicts the results for ImageNet1K under a 20 classes per task configuration.}
\vspace{-5mm}
\label{fig:spatio_angle}
\end{figure}

Figure~\ref{fig:spatio_angle} presents the gradient divergence across various baseline methods on CIFAR100 and ImageNet1K, evaluated under two different settings: 20 classes per task and the more challenging 2 classes per task. It is evident that, unlike existing baselines which generally overlook the alignment among client gradients, STAMP achieves significantly better gradient alignment. This improved alignment facilitates model updates that more effectively seek invariant aggregated gradient directions across clients for specific tasks, thereby enhancing the generalization capability of the aggregated model. This observation is consistent with the reduced global-local generalization gap demonstrated in Figure~\ref{fig:local-global}.

\subsection{Ablation Study on STAMP}
Table~\ref{tab:component-effectiveness} presents the ablation results for each component. The results demonstrate that both Spatio grAdient Matching (SAM) and Temporal grAdient Matching (TAM) consistently enhance the average classification accuracy. Notably, SAM contributes more significantly to accuracy improvement by enhancing generalization across tasks within a single communication round. In contrast, TAM plays a more critical role in reducing average forgetting by mitigating catastrophic forgetting; it achieves this by aligning the learned gradients with those from previous tasks on the same client. Additionally, the use of the prototypical coreset selection method further boosts the performance of STAMP by improving data representation through ProtoNet.
\begin{table}[h!]
\vspace{-3mm}
\centering
\caption{We conduct ablation studies on the CIFAR100 and ImageNet1K datasets, using 10 clients and 2 classes per task. Specifically, \textbf{(1)} refers to spatio-temporal gradient matching performed on the server side, \textbf{(2)} denotes temporal gradient matching executed on the client side, and \textbf{(3)} corresponds to the use of a prototypical coreset implemented with ProtoNet.}
\label{tab:component-effectiveness}
\footnotesize
\setlength\tabcolsep{3pt}
\begin{tabular}{lccccccccc}
\toprule
\textbf{Dataset} & \textbf{Metric} & \textbf{FedAvg} & \textbf{(1)} & \textbf{(2)} & \textbf{(1) + (2)} & \textbf{(1) + (3)} & \textbf{(2) + (3)} & \textbf{STAMP} \\
\midrule
\multirow{2}{*}{CIFAR100} 
& Acc. & \pmval{31.7}{1.7} & \pmval{38.1}{1.3} & \pmval{37.8}{0.6} & \pmval{44.7}{1.5} & \pmval{46.1}{0.7} & \pmval{44.9}{1.4} & \pmval{52.8}{0.9} \\
& AF & \pmval{22.1}{1.3} & \pmval{23.8}{0.4} & \pmval{21.7}{0.9} & \pmval{21.5}{1.0} & \pmval{24.7}{1.4} & \pmval{21.8}{0.6} & \pmval{24.3}{0.8} \\
\midrule
\multirow{2}{*}{ImageNet1K} 
& Acc. & \pmval{24.3}{5.1} & \pmval{30.5}{2.8} & \pmval{28.3}{2.6} & \pmval{34.1}{0.7} & \pmval{37.4}{1.1} & \pmval{36.5}{1.3} & \pmval{41.5}{2.8} \\
& AF & \pmval{19.6}{0.1} & \pmval{26.1}{0.7} & \pmval{23.8}{0.6} & \pmval{24.3}{0.9} & \pmval{26.1}{1.8} & \pmval{23.3}{0.8} & \pmval{24.2}{0.8} \\
\bottomrule
\end{tabular}
\vspace{-5mm}
\end{table}



\section{Conclusion}
In this paper, we have tackled the challenges of FCL in realistic settings characterized by client data heterogeneity and task conflicts. Recognizing the limitations of existing generative replay-based methods, we have introduced a novel model-agnostic approach, Spatio-Temporal Gradient Matching with Network-Free Prototype Coreset. Our method effectively mitigates catastrophic forgetting and data bias by leveraging prototype samples for robust gradient approximation and applying gradient matching both temporally and spatially. Through extensive experiments, we have demonstrated that our approach consistently outperforms existing baselines, highlighting its potential as a powerful solution for resilient FCL in diverse, dynamic environments.

\bibliography{neurips_2025}
\bibliographystyle{icml}

\clearpage
\appendix
\section{Detailed Algorithms}
\begin{algorithm}[H]
\caption{Pseudo Algorithm of STAMP. The \colorbox{spatio}{box} refers to the \textbf{\underline{S}}patio gr\textbf{\underline{A}}dient \textbf{\underline{M}}atching (SAM), the \colorbox{temporal}{box} refers to the \textbf{\underline{T}}emporal gr\textbf{\underline{A}}dient \textbf{\underline{M}}atching (TAM), the \colorbox{proto}{box} refers to the \textbf{\underline{p}}rototypical \textbf{\underline{c}}oreset \textbf{\underline{s}}election (PCS).}
\label{alg:FedOMG}
    \KwIn{set of source clients $\gU_\gS$, number of communication rounds $R$, local learning rate $\eta$, global learning rate $\eta_g$, searching space hyper-parameter $\kappa$.}
    \KwOut{$\theta_g^{(R)}$}

    \textbf{Clients Update:}\\
    \For{client $u\in\gU_\gS$}{
        \textbf{Receive} global model $\theta_{u}^{(r)} = \theta_g^{(r)}$\;
        \colorbox{proto}{Compute $p^l = \frac{1}{\sum^T_{t=1} \vert\gN^t_l\vert} \Big[g(\widetilde{x}^l; \phi)\cdot\sum^{t-1}_{j=1} \vert\gN^j_l\vert + \sum^{}_{i\in\gN^{t}_l} g(x_i; \phi)\Big] \cdot \mathbbm{1}\{y_j = l\},$} \\
        \colorbox{proto}{Initialize learnable coefficient set $A = \{a_i \vert i\in\gN^{t}_l\}$} \\
        \colorbox{proto}{Solve $\widetilde{X}^l = \argmin_{A} \Big\Vert \Big[\frac{1}{\vert\gM^l\vert}\sum^{}_{i\in\gM^l} g(x_i; \phi) + \frac{1}{\vert\gN^{t}_l\vert}\sum^{}_{i\in\gN^{t}_l} a_i \cdot g(x_i; \phi)\Big] - p^l \Big\Vert^2,$} \\
        \colorbox{proto}{$\widetilde{x}^l = \textrm{MixStyle}(\widetilde{x}^l; x),$} \\
        \colorbox{proto}{Save new proto into replay memory $\gM^t = \widetilde{x}^l$.}  \\
        \For{local epoch $e \in E$}{
                Sample mini-batch $\zeta$ from local data $\mathcal{D}_u$\;
                Calculate gradient $g^{t,r,e}_u = \nabla \gE(\theta^{(r,e)}_u, \zeta)$\;
            }
        \colorbox{temporal}{Calculate $\widetilde{g}^{t} = \frac{1}{E}\sum^{E}_{e=1} g^{t,r,e}_u$.} \\
        \For{task $i = 1,\ldots,t-1$}
        {
            \colorbox{temporal}{Sample coreset $\zeta$ from replay memory $\gM^i$ according to task $i$}, \\
            \colorbox{temporal}{Calculate task-wise gradients: $\widetilde{g}^i_u = \nabla \gE(\theta^{(r,e)}_u, \zeta)$}. \\    
        }
        \colorbox{temporal}{$\rvg = [\widetilde{g}^1_u, \ldots, \widetilde{g}^{t}_u]$, and $\Bar{g} = \sum^{t}_{i=1} g^i_u$}, \\
        \colorbox{temporal}{Solve: 
            $\Gamma^* = \arg\min_{\Gamma} \Gamma\rvg\cdot \Bar{g} + \kappa\Vert \Bar{g}\Vert\Vert \Gamma\rvg^{(t,r)}\Vert,$} \\
        \colorbox{temporal}{Update TAM: $g_\textrm{TAM} = \Bar{g} + \frac{\kappa\Vert \Bar{g}\Vert}{\Vert \Gamma^*\rvg^{(r)}\Vert}\Gamma^*\rvg^{(t,r)},$} \\
        \colorbox{temporal}{Model steps with aggregated gradient: 
        $\theta_{u}^{(t,r)} = \theta_{u}^{(t,r-1)} - \eta_g g^{(t,r)}_\textrm{TAM}.$} \\
        Upload client's model $\theta^{(t,r+1)}_u$ to server\;    
    }
    \textbf{Server Optimization:}\\
    \For{task $t = 0 ,\ldots$}{
        \For{round $r=0,\ldots, R$}{
            \textbf{Clients Updates}\;
            Calculate $g^{(t,r)}_{u} = \theta^{(t,r+1)}_u - \theta^{(t,r)}_u$, $\rvg^{(t,r)} = \{g^{(t,r)}_{u} \vert u\in\gU_\gS\}$\;
            Calculate $g^{(t,r)}_{FL}$ (e.g., $g^{(t,r)}_{FL} = \frac{1}{U} \sum_{u=1}^{U} g^{(t,r)}_{u}$ as the FedAvg update)\;
            \colorbox{spatio}{Solve: 
                $\Gamma^* = \arg\min_{\Gamma} \Gamma\rvg^{(t,r)}\cdot g^{(t,r)}_{\textrm{FL}} + \kappa\Vert g^{(t,r)}_{\textrm{FL}}\Vert\Vert \Gamma\rvg^{(t,r)}\Vert,$} \\
            \colorbox{spatio}{Update SAM: $g^{(t,r)}_\textrm{SAM} = g^{(t,r)}_{\textrm{FL}} + \frac{\kappa\Vert g^{(t,r)}_{\textrm{FL}}\Vert}{\Vert \Gamma^*\rvg^{(t,r)}\Vert}\Gamma^*\rvg^{(t,r)},$} \\
            \colorbox{spatio}{Model steps with aggregated gradient: 
            $\theta_u^{(t, r+1)} = \theta_{u}^{(t,r)} - \eta_u g^{(t,r)}_\textrm{SAM}.$}
        }
    }
\end{algorithm}

\clearpage
\section{Related Works}
\subsection{Importance-based Sampling}
LGA \citep{10323204} introduces a method to balance the contributions of different classes to the gradient, aiming to mitigate catastrophic forgetting caused by imbalance among incremental tasks.
Re-Fed \citep{Li_2024_CVPR} presents a method for quantifying an importance score, which is utilized to selectively retain cached samples within the replay memory.
FedWeIT \citep{2021-FCL-FedWeIT} partitions network weights into global federated and sparse task-specific parameters, enabling clients to selectively acquire knowledge through a weighted combination of others' task-specific parameters.
FedSSI \citep{2025-FCL-FedSSI} introduces a regularization technique that estimates the importance of each synaptic weight change during training. It penalizes substantial changes to weights deemed important for previously learned tasks, thereby helping to preserve prior knowledge.
\subsection{Prototype-based Learning}
SR-FDIL \citep{10620614} introduces an approach that utilizes data from the local replay memory to train both the prototype generator and the discriminator on local devices.
TagFed \citep{10658168} proposes a method to identify repetitive data features from previous tasks and augment them for the current task prior to federation, thereby enhancing overall performance.
\subsection{Gradient Memory}
GradMA \citep{Luo_2023_CVPR} employs gradient projection on the client side, correcting gradients via quadrature optimization using stored gradients from other clients.
\subsection{Generative Replay Memory}
FedCIL \citep{2023-FCL-FedCIL} introduces an efficient approach for training GAN-based replay memory in distributed systems.
TARGET \citep{2023-FCL-TARGET} introduces an approach that learns a server-side generative model capable of producing data that adheres to the global model distribution. This generated data is subsequently used to update the client-side student model via knowledge distillation.
AF-FCL \citep{wuerkaixi2024accurate} introduces a generative model that employs a learned normalizing flow to capture and retain the essential data distribution while effectively eliminating biased features.
pFedDIL \citep{10.1007/978-3-031-72952-2_8} proposes an approach that transfers knowledge across incremental tasks by using a small auxiliary classifier in each personalized model to distinguish its specific task from others.
FBL \citep{Dong_2023_CVPR} uses adaptive class-balanced pseudo labeling along with semantic compensation and relation consistency losses to generate reliable pseudo labels and balance gradient propagation, thereby mitigating the effects of background shifts.

\subsection{Episodic Replay Memory for Continual Learning}
GEM \citep{NIPS2017_f8752278} introduced an episodic memory mechanism that stores a subset of data samples, enabling the estimation of task-specific gradients. This approach facilitates gradient projection, thereby mitigating catastrophic forgetting in CL. VR-MCL \citep{wu2024meta} introduced a meta CL approach that effectively utilizes data stored in the memory buffer. 

Authors in \citep{2023-FCL-FedCIL} demonstrate that incorporating a GAN-based replay memory in a distributed system can be significantly affected by feature shifts among clients. To address this challenge, FedCIL introduces a distillation-based approach designed to mitigate discrepancies across different domains.
GPM \citep{saha2021gradient} introduces a method for storing gradient projections in replay memory as an alternative to retaining previous data, thereby facilitating CL. FS-DGPM \citep{deng2021flattening} introduces an enhanced version of GPM, in which the projected gradients are flattened. This flattening process improves generalization and enhances robustness to noise caused by a sharp loss landscape.

\clearpage
\section{Experimental Details}\label{app:settings}
We utilize the pFLLib framework \citep{zhang2025pfllib} as FL core framework to design the FCL settings. All experiments are conducted using four NVIDIA GeForce RTX 4090 GPUs and two NVIDIA GeForce RTX 3090 GPUs. The detailed experimental configurations are outlined below:
\subsection{Datasets}

\subsubsection{Heterogeneous Federated Continual Learning Settings}

Our work investigates the behavior of various algorithms in a heterogeneous FCL setting. To align with a realistic and challenging non-IID federated scenario, we increase the difficulty by adopting the task design proposed by \citep{Dohare2024}, in which we construct a sequence of classification tasks by taking the classes in groups. 
\begin{example}
    For example, in case of binary classification, one task could involve differentiating chickens from llamas, while another might focus on differentiating phones from computers.
\end{example}

To consider the performance of baselines under different level of heterogeneity, we consider two experimental scenarios. In the first, each task comprises $20$ distinct classes. This setup represents the conventional task configuration commonly used in existing literature \citep{wuerkaixi2024accurate}. In the second, each task contains only $2$ classes, creating a more challenging environment. In this case, models are more likely to overfit to individual tasks, making them more susceptible to catastrophic forgetting when adapting to new tasks. Simultaneously, client divergence becomes more pronounced under this configuration.

Specifically, we utilize two widely adopted benchmark datasets:

\textbf{Non-Overlapped-CIFAR100.}
The CIFAR100 dataset \citep{krizhevsky2009learning} consists of 100 object
categories, with a total of 60,000 images. Each image has a resolution of $32 \times 32$ pixels. In case 1 task comprises 2 classes, we can form 4950 distinct tasks. In case 1 task comprises 20 classes, we can form more than $5\!\!\times\!\!10^{20}$ distinct tasks.

\textbf{Non-Overlapped-ImageNet1K.}
ImageNet1K dataset \citep{5206848} contains 1,000 diverse object categories, with over 1.3 million high-resolution training images. All images are resized to $224 \times 224$ pixels during preprocessing. In case 1 task comprises 2 classes, we can form half a million tasks. In case 1 task comprises 20 classes, we can form more than $3\!\!\times\!\!10^{41}$ distinct tasks. The scale and diversity of ImageNet1K pose greater challenges in terms of memory footprint, computational cost, and model scalability.

\subsection{Baselines}
We evaluate our approach against several established baselines from FL, and FCL. For conventional FL baselines, we compare with standard methods such as FedAvg \citep{2017-FL-FedAvg}, FedALA \citep{2023-FL-FedALA}, FedDBE \citep{2023-FL-FedDBE}, FedL2P \citep{2023-FCL-FedL2P}, and FedAS \citep{2024-FL-FedAS}, FedOMG \citep{2025-FDG-FedOMG}. 
FedAvg serves as the foundational baseline in FL. FedALA, FedL2P, and FedAS focus on personalized FL, enabling models to adapt to client-specific tasks and thereby mitigating the effects of task heterogeneity. In contrast, FedDBE and FedOMG aim to construct a more robust global model by reducing inter-client bias, thereby enhancing generalization across both tasks and clients.

For FCL, we assess several state-of-the-arts, including FedWeIT \citep{2021-FCL-FedWeIT}, GLFC \citep{dong2022federated}, FedCIL \citep{2023-FCL-FedCIL}, LANDER \citep{2024-FCL-Lander}, TARGET \citep{2023-FCL-TARGET}, FedSSI \citep{2025-FCL-FedSSI}, and AF-FCL \citep{wuerkaixi2024accurate}. FedWeIT exemplifies approaches that allocate specialized expert modules for each task, allowing task-specific adaptation. GLFC uses a distillation-based approach to address catastrophic forgetting, considering both local and global aspects. FedCIL, LANDER, TARGET, and AF-FCL adopt generative replay strategies, training generative models on each client to synthesize pseudo-data for previously encountered tasks. Among these, AF-FCL is the most recent and directly addresses the challenges posed by heterogeneous FCL settings, making it a particularly relevant benchmark for comparison. 

\subsection{Evaluation Metrics}
To evaluate the baselines, we utilize two standard metrics from the CL literature \citep{2021-FCL-FedWeIT}, \citep{mirzadeh2021linear}, which are well-suited for tracking the performance of a global model in FL, coined accuracy and averaged forgetting.


\textbf{Averaged Forgetting.}
This metric measures the decline from a task’s highest accuracy, which is typically achieved right after it is trained, to its final accuracy after all tasks have been learned. For $T$ tasks, the forgetting is defined as
\begin{align}
    AF = \frac{1}{T-1} \sum_{i=1}^{T-1} \max_{t \in {1,\ldots,T-1}} (a_{t,i} - a_{T,i}).
\end{align}
As the model shifts focus to new tasks, its performance on earlier ones often decreases. Therefore, minimizing forgetting is important to maintain overall performance.

\subsection{Architecture Details}
For CIFAR-10, CIFAR100, Digit10, and Office31, we adopt conventional ResNet-18 \citep{he2016deep} as the backbone network architecture for all validation experiments. For ImageNet1K, we employ Swin Transformer Tiny (Swin-T) \citep{liu2021swin} as the backbone. It is noted that FCIL, LANDER, TARGET, FedL2P, FedWeIT and AF-FCL use addition generative networks or modify their network architectures, with details summarized in the following table. We denote FedWeIT (T) as the version theoretically proposed in the original paper, while FedWeIT (C) represents the configuration observed in our experimental implementation.

\begin{table}[h]
\centering
\caption{Architectural details of methods with modified models.}
\label{tab:modified-arch}
\begin{tabular}{l|cc|cc}
\toprule
\textbf{Method} & \multicolumn{2}{c|}{\textbf{CIFAR-10, CIFAR100, Digit10, Office31}} & \multicolumn{2}{c}{\textbf{ImageNet1K}} \\
\midrule
& \textbf{Model} & \textbf{\#Params} & \textbf{Model} & \textbf{\#Params} \\
\midrule
FedAvg & ResNet-18 & 11.7 M & Swin-T & 28.8 M \\
FedSSI & ResNet-18 & 11.7 M & Swin-T & 28.8 M \\
\midrule
FCIL & ResNet-18 + GAN & 16.1 M & Swin-T + GAN & 49.7 M \\
LANDER & ResNet-18 + GAN & 16.1 M & Swin-T + GAN & 49.7 M \\
TARGET & ResNet-18 + GAN & 16.1 M & Swin-T + GAN & 49.7 M \\
FedL2P & ResNet-18 + Meta-Net & 13.5 M & Swin-T + Meta-Net & 32.6 M \\
\midrule
FedWeIT (T) & Modified ResNet-18 & 596.2 M & Modified Swin-T & 7192.3 M \\
FedWeIT (C) & Modified LeNet & 171.8 B &  & \\
AF-FCL & ResNet-18 + NFlow & 21.3 M & Swin-T + NFlow & 53.4 M \\
\bottomrule
\end{tabular}
\end{table}
Specifically, FedWeIT augments the base model with sparse task-adaptive parameters, task-specific masks over local base parameters, and attention weights for inter-client knowledge transfer. FCIL, LANDER, and TARGET incorporate additional GANs to learn past task features. FedL2P introduces a meta-net that generates personalized hyper-parameters, such as batch normalization statistics and learning rates, adapted to each client's local data distribution to improve learning on non-IID data. AF-FCL additionally requires a normalizing flow generative model (NFlow\footnote[1]{NFlow refers to the normalizing flow model, where the example is provided in \url{https://github.com/zaocan666/AF-FCL/blob/main/FLAlgorithms/PreciseFCLNet/model.py}}) for credibility estimation and generative replay mechanism, which guide selective retention and forgetting. 

\subsection{Training Details}
In our proposed heterogeneous federated continual learning framework for the CIFAR100 and ImageNet1K datasets, we consider a setting involving 10 clients with a client participation fraction of 1.0. We do not adopt a conventional non-IID distribution in this scenario; instead, each client is assigned distinct classes, which introduces a level of heterogeneity that is more challenging than typical non-IID configurations.

Additionally, we evaluate the proposed approach under non-IID conditions using four benchmark datasets: CIFAR-10, CIFAR100, Digit-10, and Office-31. For these experiments, we simulate data heterogeneity using the Dirichlet distribution with varying concentration parameters (e.g., $\alpha = 0.1$, $1.0$, $10.0$, and $100.0$) to control the degree of non-IID-ness. The complete details of the experimental settings are provided in Table~\ref{tab:training-details}.

\begin{table}[ht]
\centering
\caption{Experimental Details. Settings for heterogeneous and non-IID distributed FCL.}
\label{tab:training-details}
\resizebox{\textwidth}{!}{%
\begin{tabular}{lcc|cccc}
\toprule
\multirow{2}{*}{\textbf{Attributes}} & \multicolumn{2}{c|}{\textbf{Heterogeneous FCL}} & \multicolumn{4}{c}{\textbf{Non-IID distributed FCL}} \\
\cmidrule(lr){2-3} \cmidrule(lr){4-7}
 & \textbf{CIFAR100} & \textbf{ImageNet1K} & \textbf{CIFAR10} & \textbf{CIFAR100} & \textbf{Digit10} & \textbf{Office31} \\
\midrule
Task size & 141 MB / 14 MB & 8 GB / 0.8 GB & 141 MB & 141 MB & 480 M & 88 M \\
Image number & 60K & 1.3M & 60K & 60K & 110K & 4.6K \\
Image Size & $3 \times 32 \times 32$ & $3 \times 224 \times 224$ & $3 \times 32 \times 32$ & $3 \times 32 \times 32$ & $1 \times 28 \times 28$ & $3 \times 300 \times 300$ \\
Task number & 5 / 50 & 50 / 500 & 5 & 10 & 4 & 3 \\
\midrule
Batch Size & 128 & 128 & 64 & 64 & 64 & 32 \\
Learning Rate & 0.005 & 0.005 & 0.01 & 0.01 & 0.001 & 0.01 \\
Data heterogeneity & N/A & N/A & 0.1 & 10.0 & 0.1 & 1.0 \\
Client numbers & 10 & 10 & 10 & 10 & 10 & 10 \\
Local training epoch & 5 & 5 & 5 & 5 & 5 & 5 \\
Client selection ratio & 1.0 & 1.0 & 1.0 & 1.0 & 1.0 & 1.0 \\
Rounds per Task & 25 & 25 & 80 & 100 & 60 & 60 \\
\bottomrule
\end{tabular}%
}
\end{table}

\section{Additional Experimental Evaluations}\label{app:add-exp}
\subsection{Experimental Evaluations on Pretrained Models}
\begin{figure}[ht]
\centering
\subfloat{\includegraphics[width=0.7\linewidth]{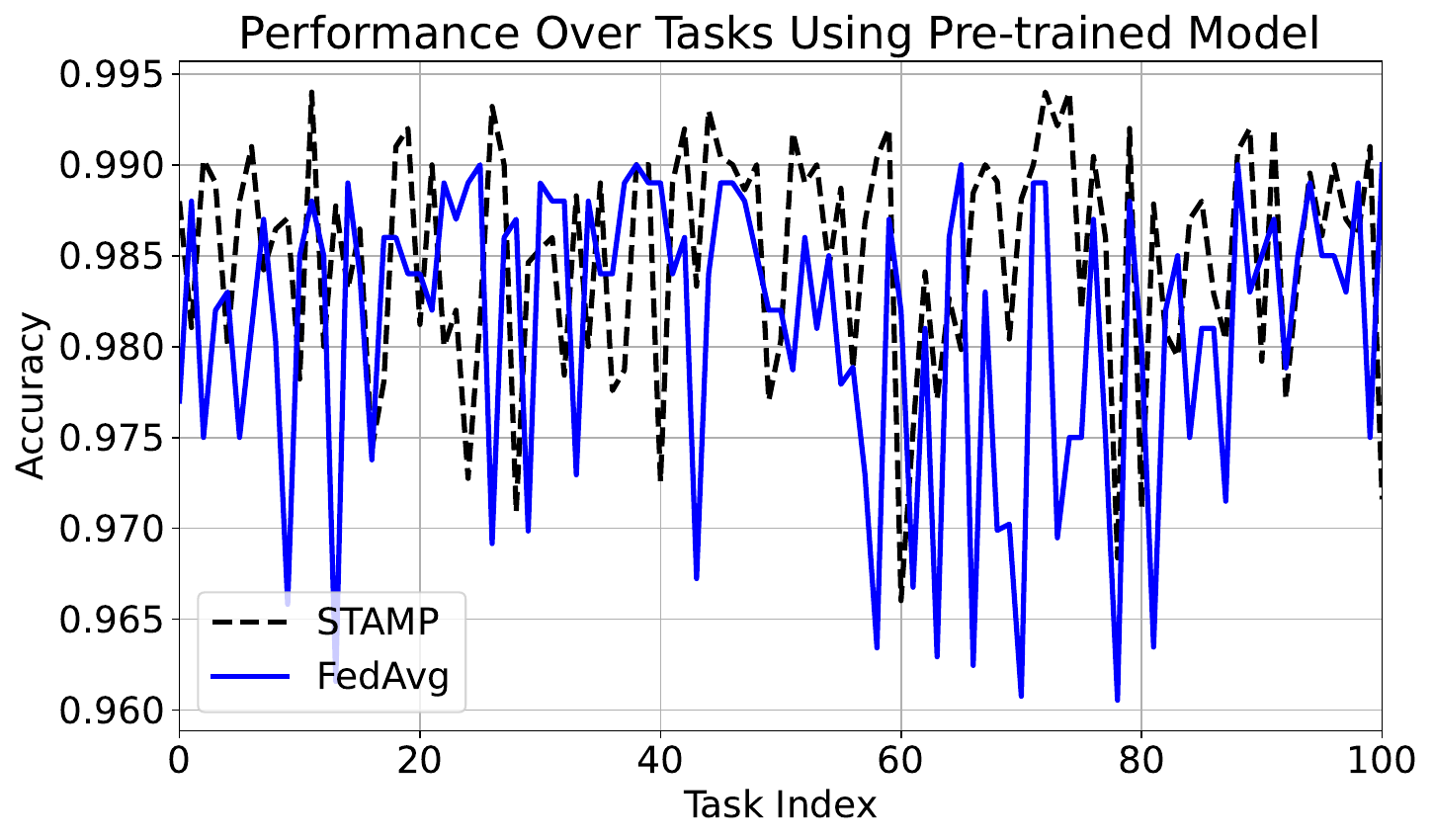}}
\caption{Accuracy on ImageNet1K with Pretrained Models.}
\vspace{-5mm}
\label{fig:ImageNet-Pretrain}
\end{figure}

Figure~\ref{fig:ImageNet-Pretrain} illustrates the performance of FedAvg and STAMP on the ImageNet1K dataset using a pretrained model. Given that the model is pretrained on the same dataset, the evaluation may suffer from overfitting. Consequently, the experimental results show no substantial performance difference between the two algorithms. Moreover, the issue of catastrophic forgetting appears to be minimal in this evaluation setting.

\clearpage
\subsection{Hyper-parameter tuning for STAMP}
In this section, we examine the impact of various hyperparameters through a series of experiments conducted on the ImageNet-1K dataset. For each experiment, one specific hyperparameter is varied while all other hyperparameters are held constant.
\subsubsection{Gradient Normalization}
\begin{figure}[ht]
\centering
\subfloat{\includegraphics[width=0.6\linewidth]{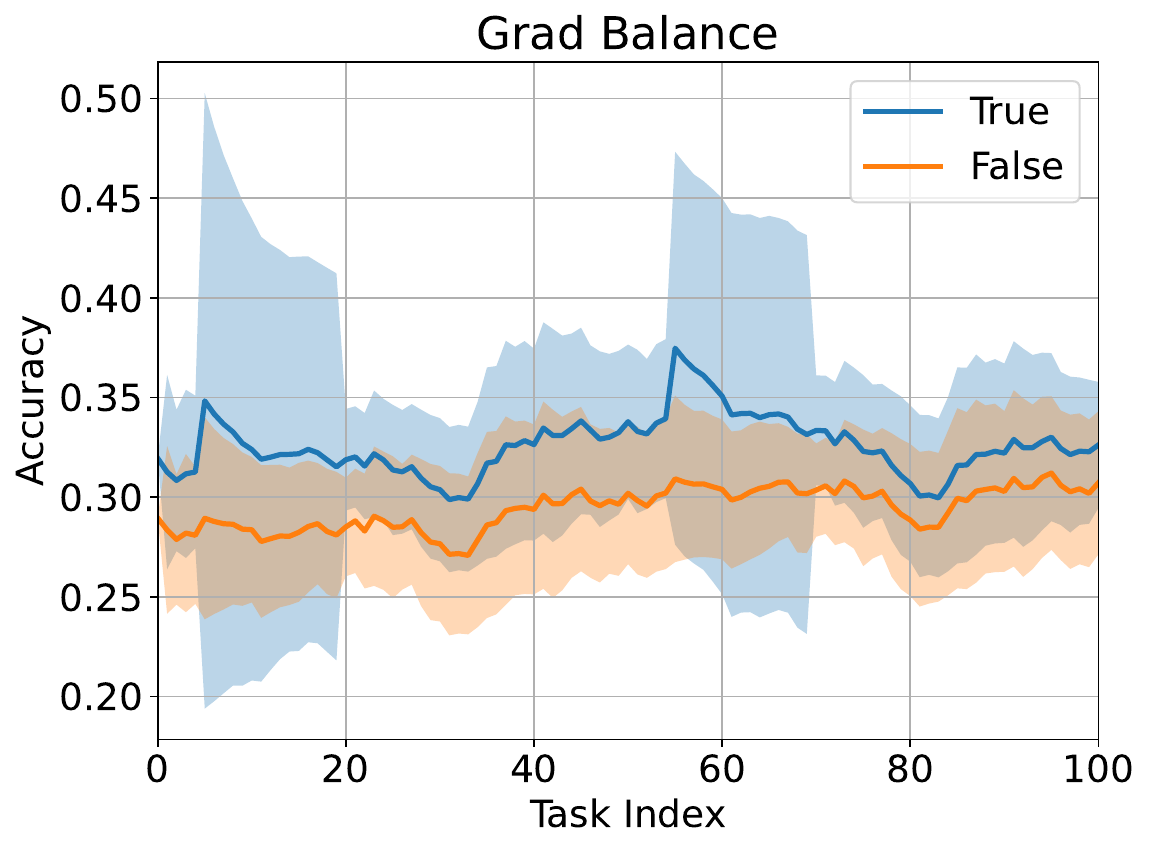}}\;
\caption{Analysis on Gradient Normalization.}
\label{fig:grad-norm}
\end{figure}

Since STAMP is sensitive to the magnitude of local gradients, the presence of a dominant subset with disproportionately large gradient magnitudes can bias the optimization process toward that subset during gradient matching. Figure~\ref{fig:grad-norm} illustrates the impact of applying gradient normalization on both the client and server sides before performing gradient matching. With gradient normalization in place, STAMP demonstrates a notable improvement in performance.

\subsubsection{Local Epoch}
\begin{figure}[ht]
\centering
\subfloat{\includegraphics[width=0.6\linewidth]{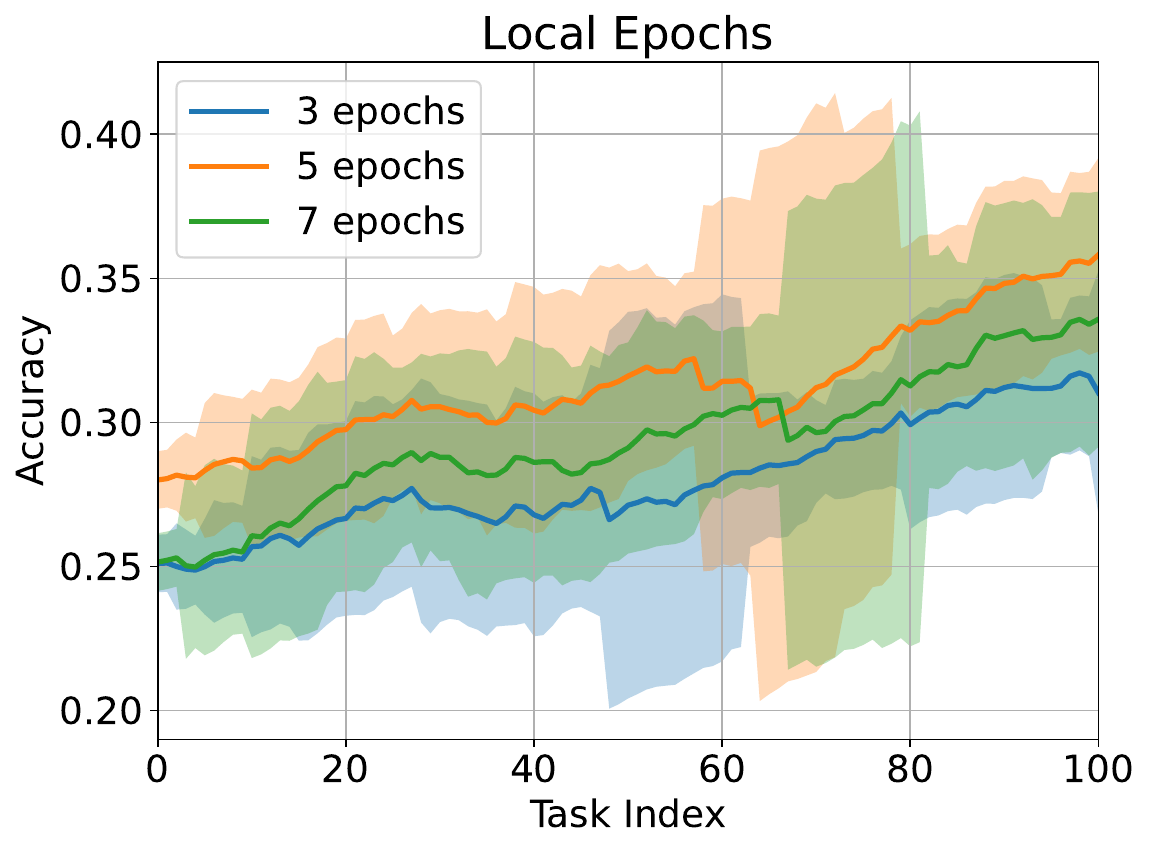}}\;
\caption{Analysis on different number of local epochs.}
\label{fig:local-epochs}
\end{figure}

Selecting the number of local epochs is crucial, as increasing the number of local epochs leads to a more accurate approximation of the local gradient trajectory. Figure~\ref{fig:local-epochs} illustrates the performance of STAMP under varying numbers of local epochs.

\newpage
\subsubsection{Local Learning Rate}
\begin{figure}[ht]
\centering
\subfloat{\includegraphics[width=0.6\linewidth]{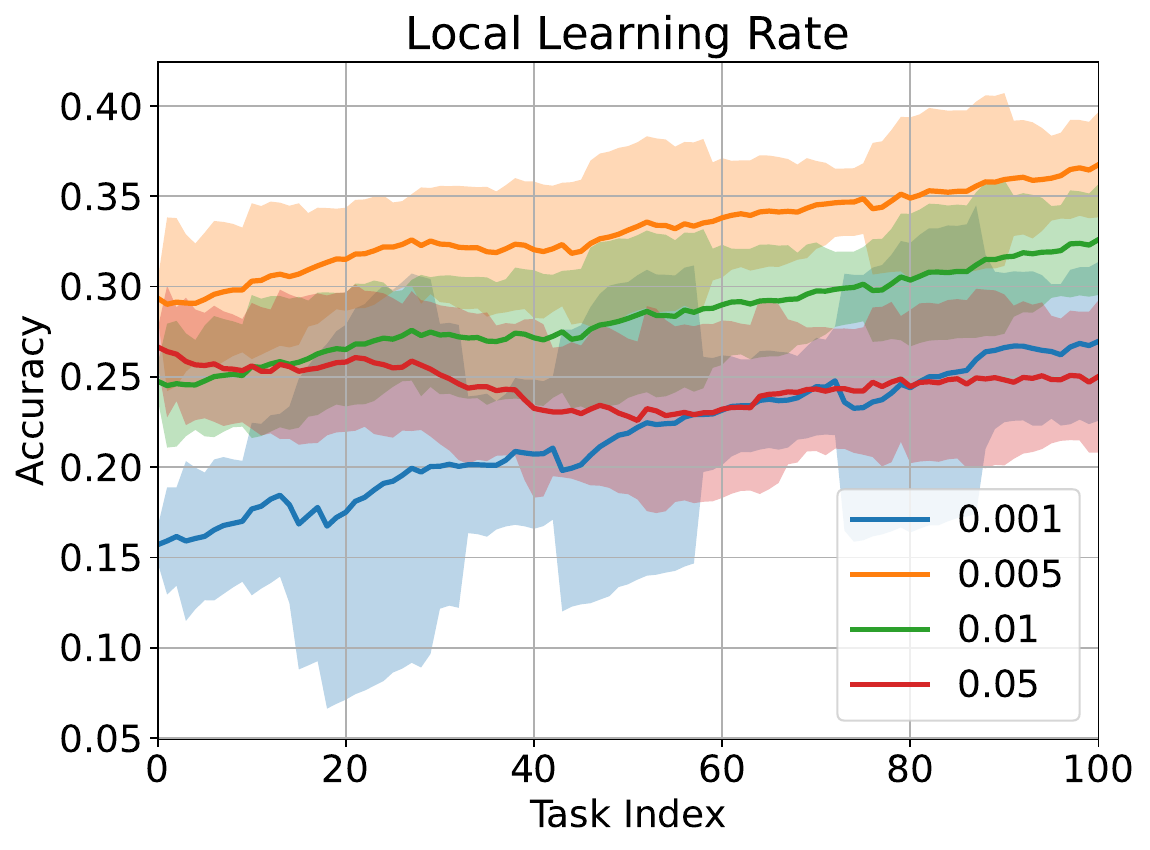}}\;
\caption{Analysis on different local learning rate.}
\label{fig:local_lr}
\end{figure}

Figure~\ref{fig:local_lr} illustrates the performance of STAMP under different local learning rate. 

\subsubsection{Gradient Matching Searching Radius}
\begin{figure}[ht]
\centering
\subfloat{\includegraphics[width=0.6\linewidth]{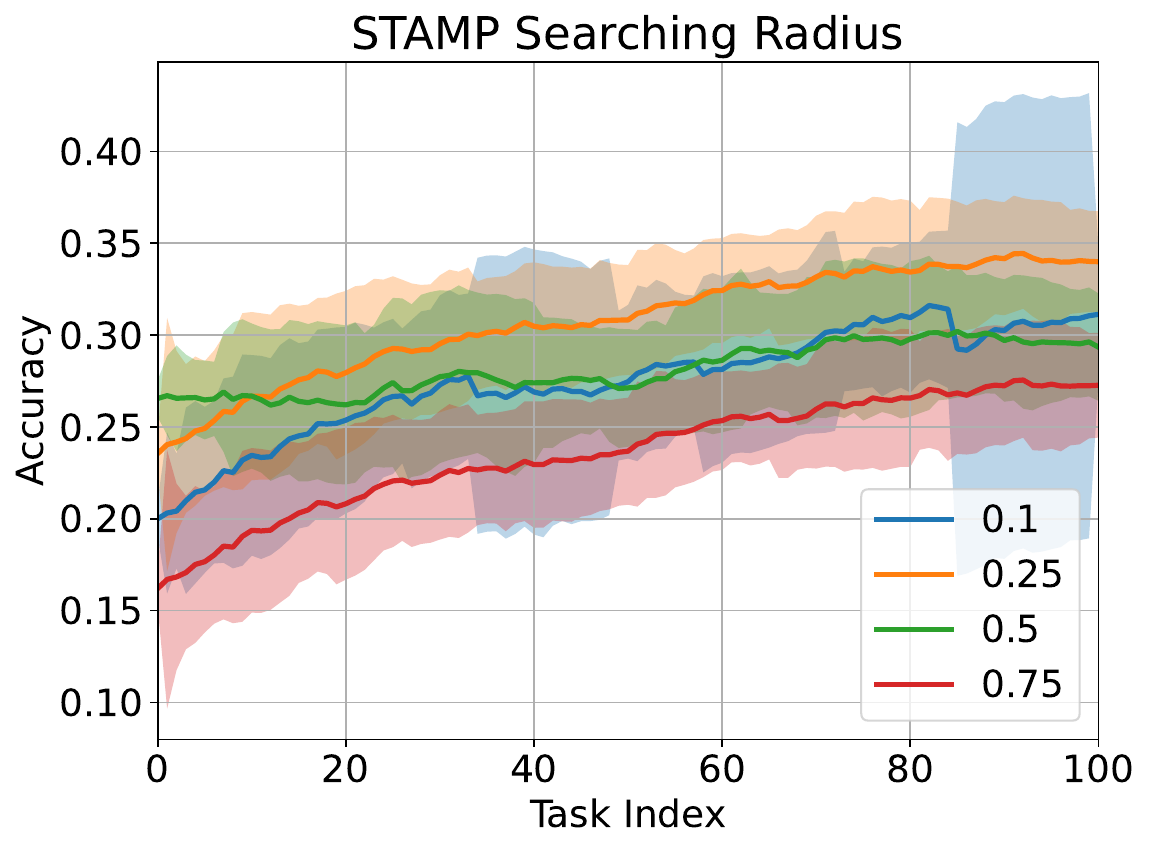}}\;
\caption{Analysis on different searching radius.}
\label{fig:stamp_c}
\end{figure}

Figure~\ref{fig:stamp_c} illustrates the impact of the search radius on gradient matching in STAMP. Selecting an appropriate search radius (e.g., $0.5$) is critical for achieving an optimal gradient matching solution. A smaller radius (e.g., $0.1$) constrains the search space too tightly, causing the solution to converge toward the average gradient and reducing matching effectiveness. Conversely, a larger radius (e.g., $0.75$) broadens the search space excessively, making it difficult to identify an optimal solution.

\newpage
\subsubsection{Gradient Matching Step Size $\&$ Momentum}
\begin{figure}[ht]
\centering
\subfloat{\includegraphics[width=0.6\linewidth]{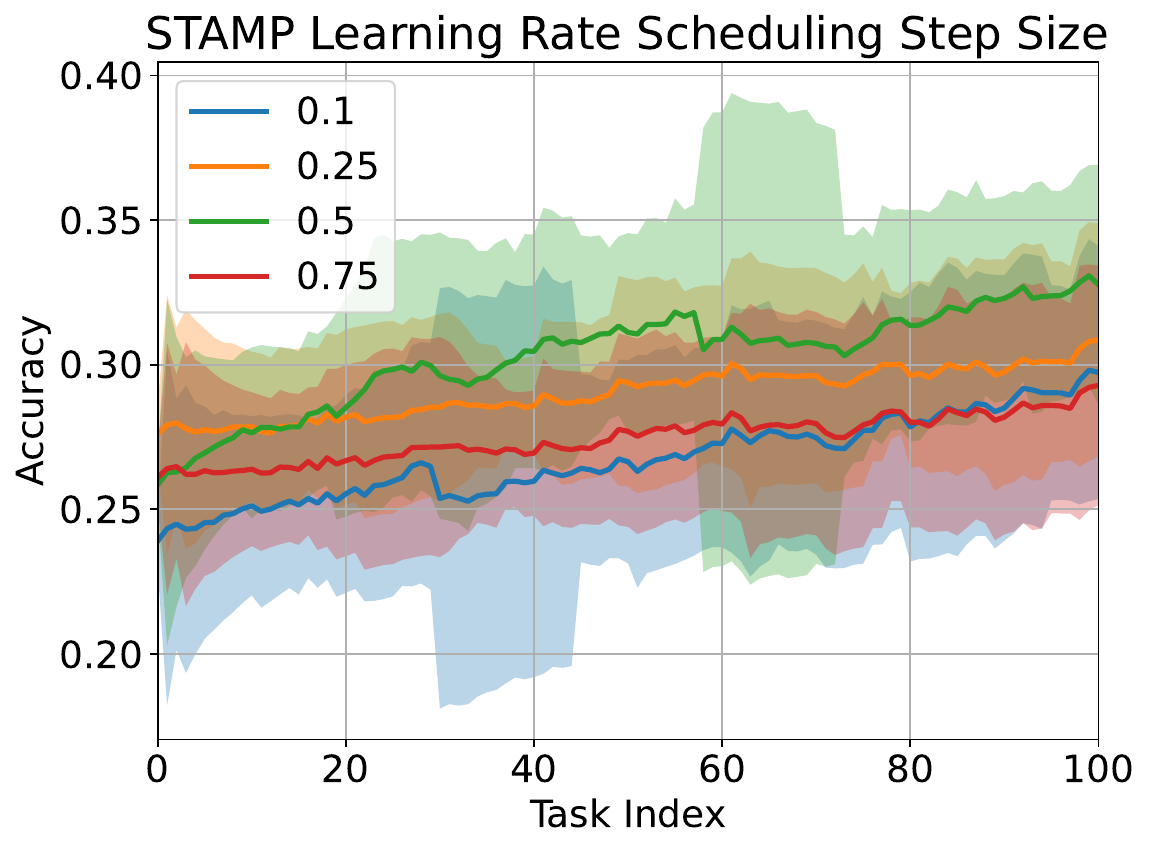}}\;
\caption{Analysis on different learning rate scheduling step size.}
\label{fig:stamp_gamma}
\end{figure}

\begin{figure}[ht]
\centering
\subfloat{\includegraphics[width=0.6\linewidth]{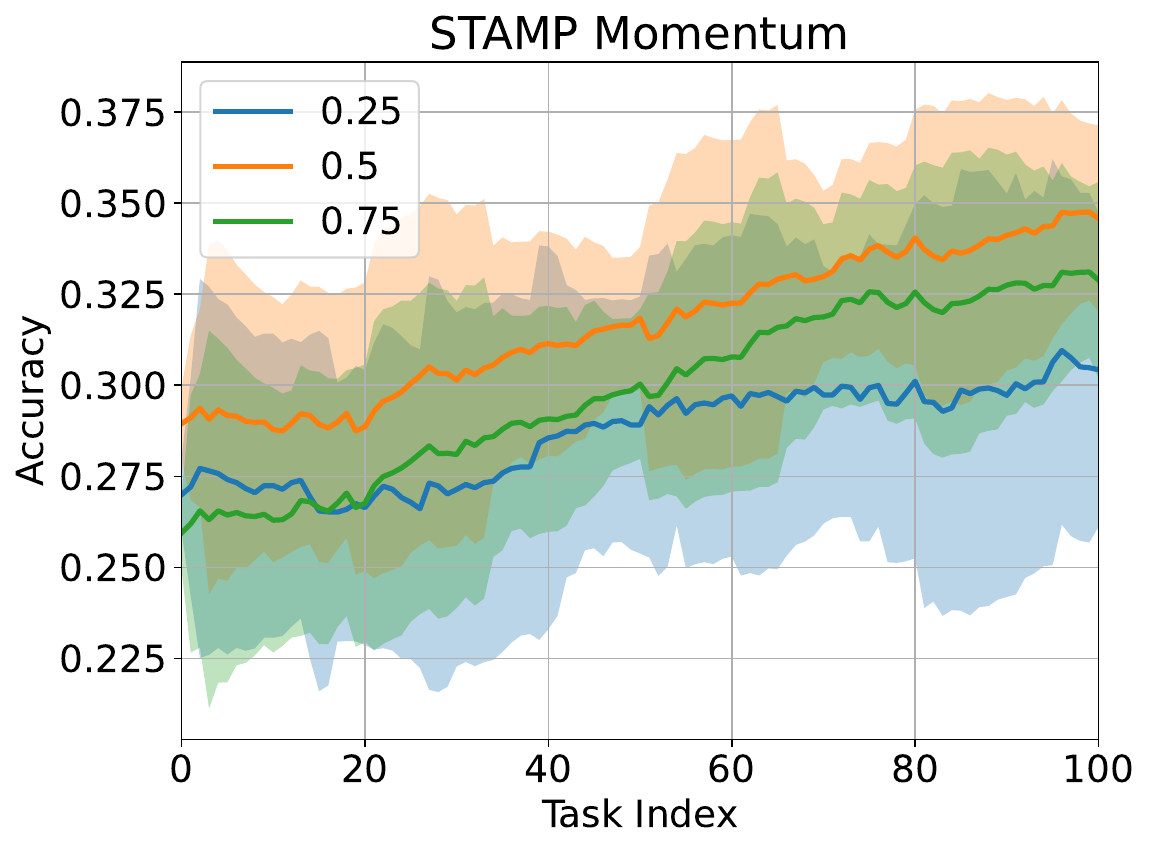}}\;
\caption{Analysis on different momentum for gradient matching.}
\label{fig:stamp_momentum}
\end{figure}

Figures~\ref{fig:stamp_gamma} and \ref{fig:stamp_momentum} demonstrate the effects of momentum and learning rate scheduling on gradient matching performance.

\newpage
\subsubsection{Gradient Matching Number of Rounds}
\begin{figure}[ht]
\centering
\subfloat{\includegraphics[width=0.6\linewidth]{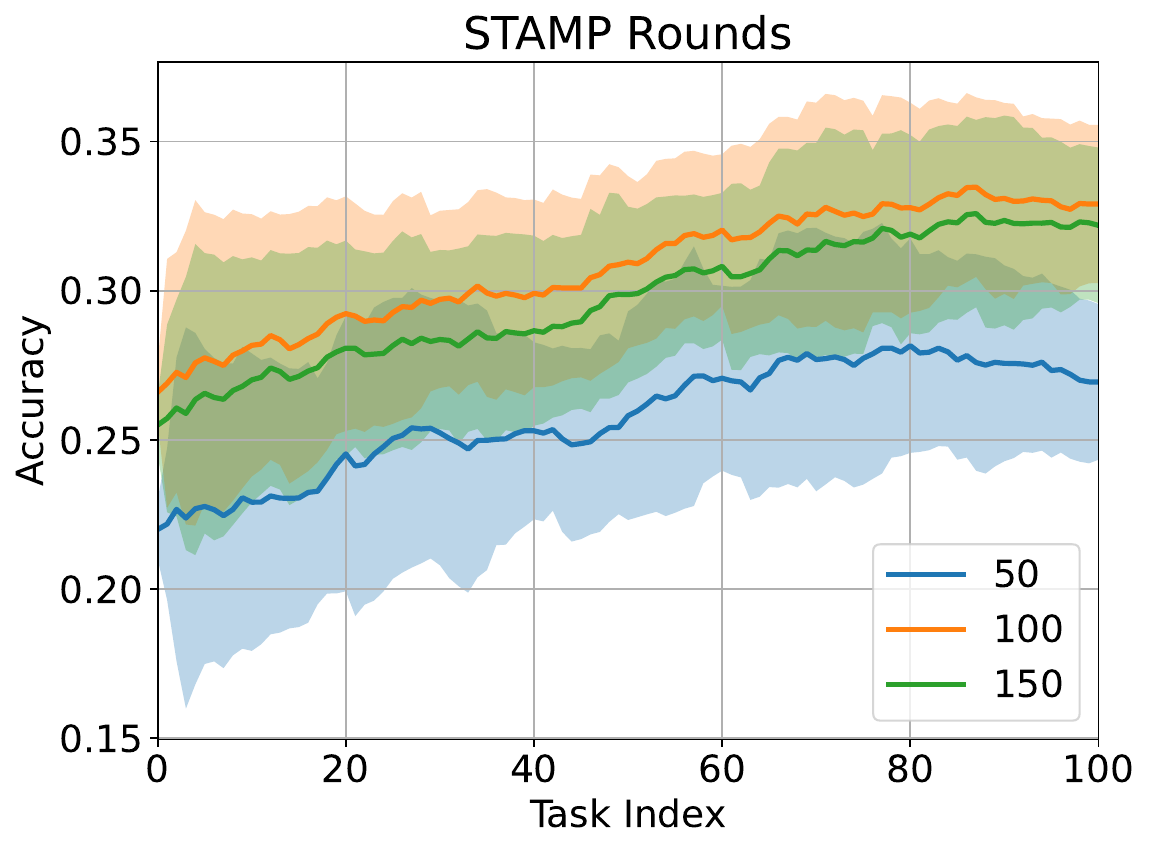}}\;
\caption{Analysis on different number of rounds}
\label{fig:stamp_round}
\end{figure}

Figure~\ref{fig:stamp_round} illustrates the impact of the number of optimization steps on the efficiency of gradient matching. 

\subsubsection{Gradient Matching Scheduling Step Size}
\begin{figure}[ht]
\centering
\subfloat{\includegraphics[width=0.6\linewidth]{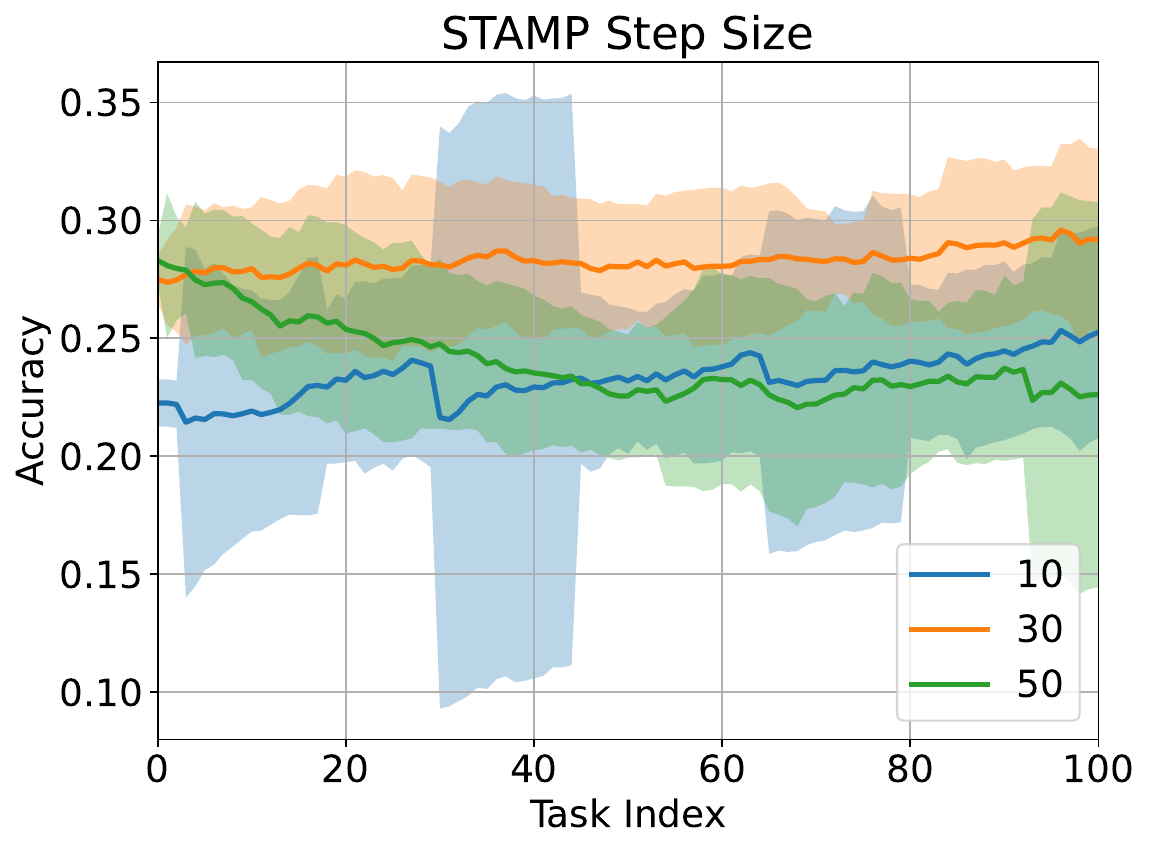}}\;
\caption{Analysis on different scheduling step size.}
\label{fig:stamp_step_size}
\end{figure}

Figure~\ref{fig:stamp_step_size} illustrates the performance of STAMP under various learning rate scheduler step sizes. Selecting an appropriate step size (e.g., $30$) facilitates optimal gradient matching decisions, thereby enhancing the stability and efficiency of FCL training.

\newpage
\subsubsection{Gradient Matching Learning Rate}
\begin{figure}[ht]
\centering
\subfloat{\includegraphics[width=0.6\linewidth]{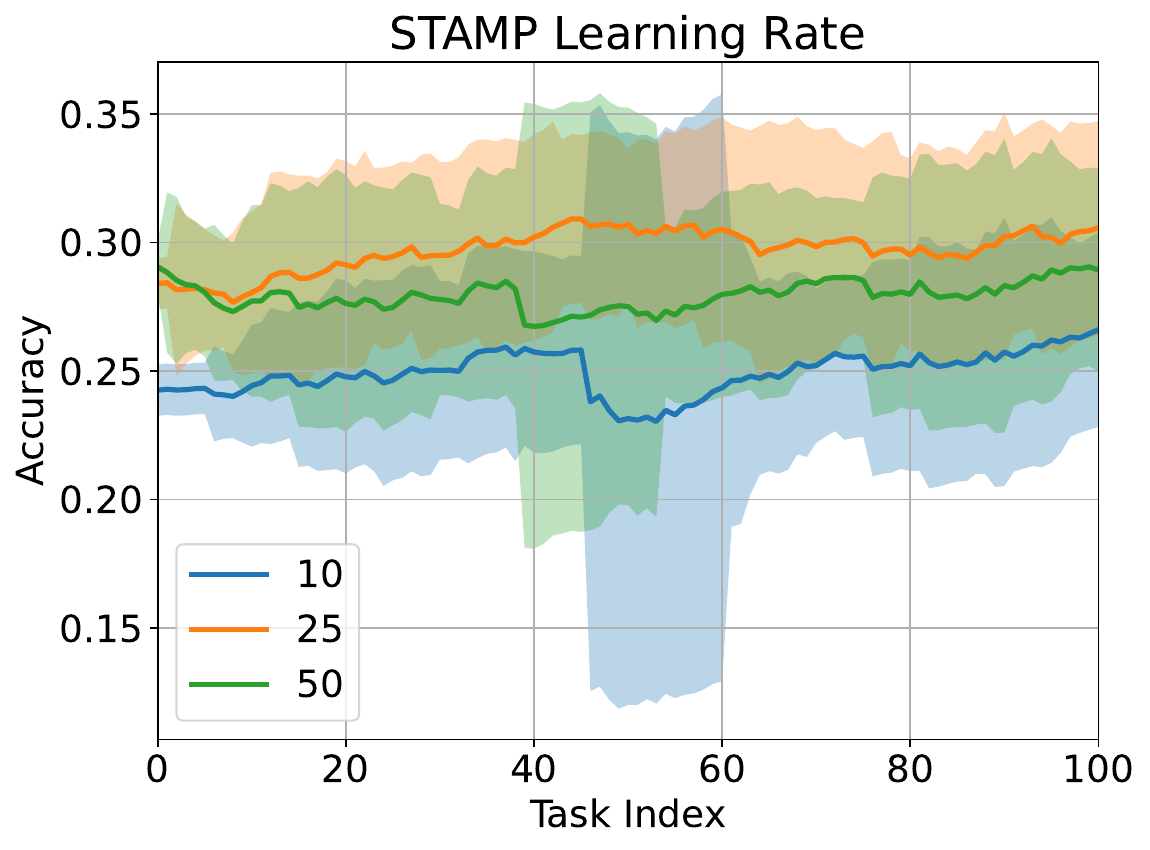}}\;
\caption{Analysis on different gradient matching learning rate.}
\label{fig:stamp_lr}
\end{figure}

Figure~\ref{fig:stamp_lr} illustrates the effect of varying learning rates on the optimization of gradient matching. The results indicate that STAMP achieves optimal performance when the learning rate is set to $25$.

\subsubsection{Global Update Learning Rate}
\begin{figure}[ht]
\centering
\subfloat[\label{fig:global_lr}]{\includegraphics[width=0.46\linewidth]{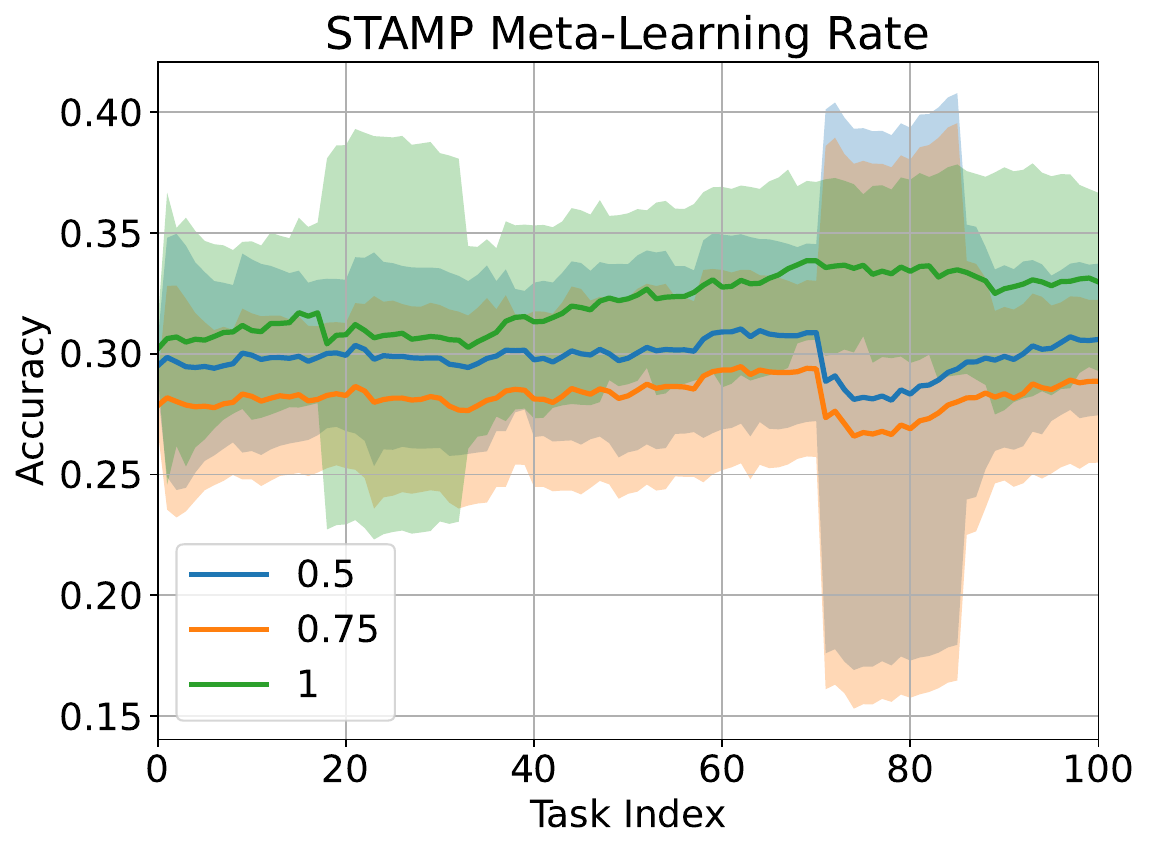}}\;
\subfloat[\label{fig:gradient_norm}]{\includegraphics[width=0.46\linewidth]{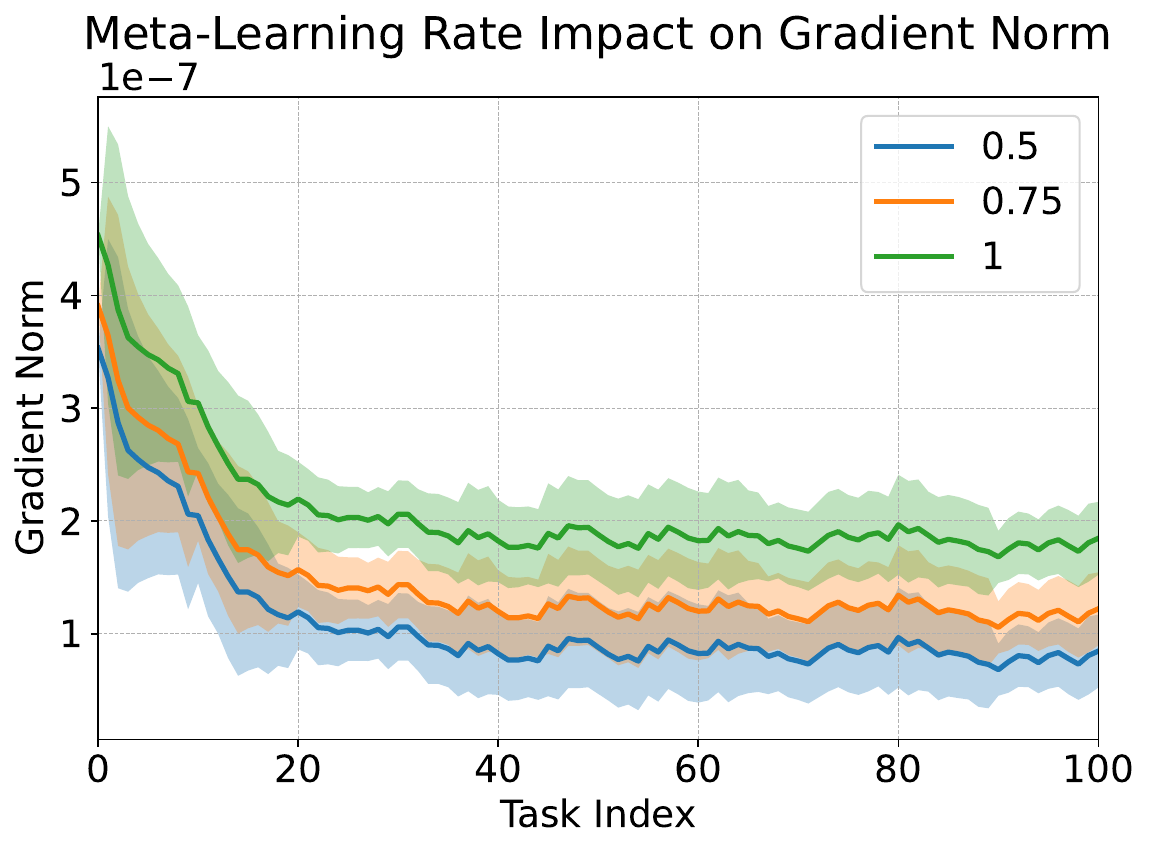}}\;
\caption{Analysis on global learning rate.}
\end{figure}

The global update learning rate significantly influences the norm of the aggregated gradient. As shown in Figure~\ref{fig:global_lr}, selecting a lower learning rate can reduce the norm of the aggregated gradient (see Figure~\ref{fig:gradient_norm}). This reduction may lead to slower convergence or result in gradient magnitudes that are insufficient to escape sharp minima.

\clearpage
\subsection{Effectiveness of Prototypical Coreset}
\begin{figure}[ht]
\centering
\subfloat{\includegraphics[width=0.7\linewidth]{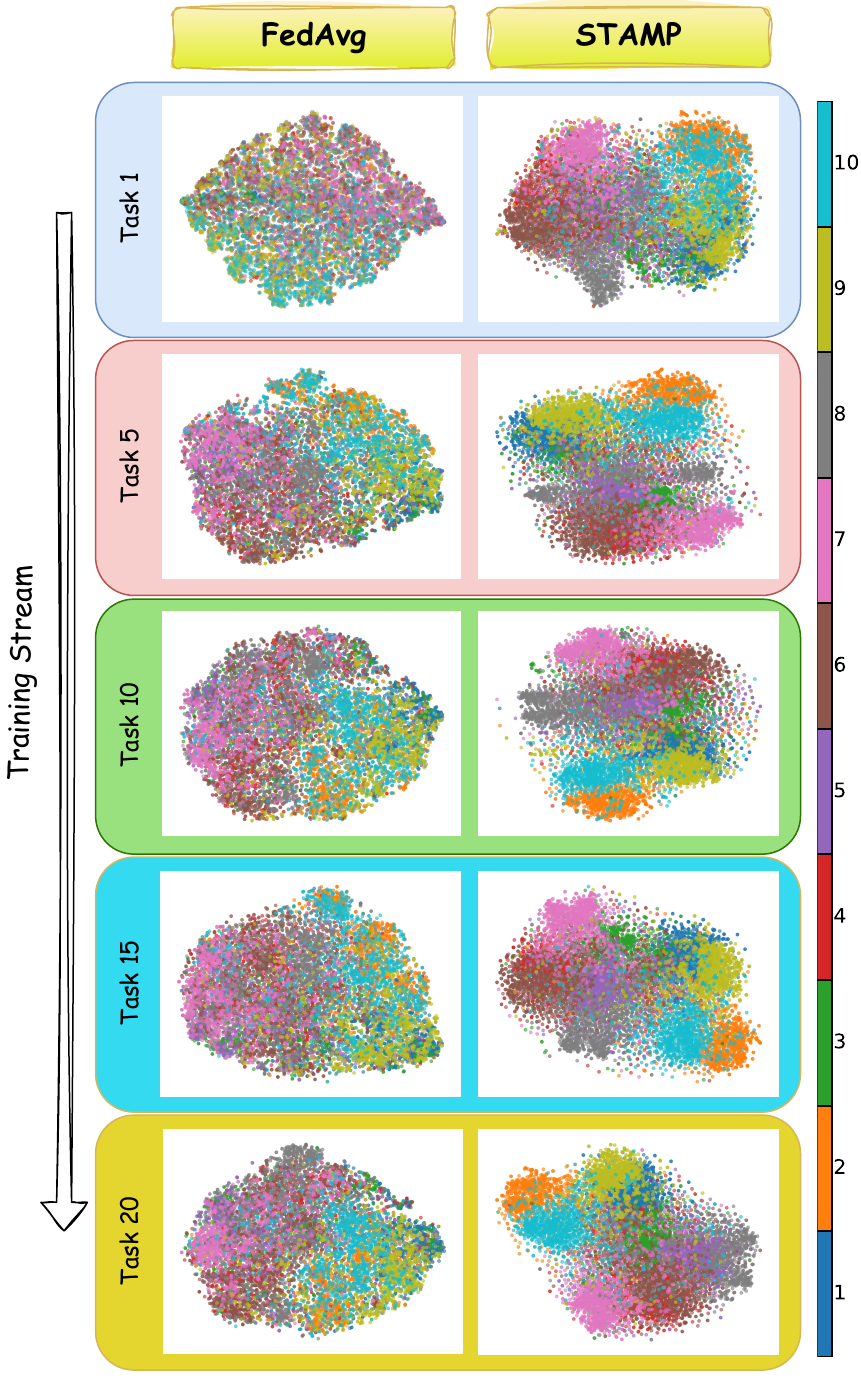}}\;
\caption{t-SNE visualizations of features learned by FedAvg and STAMP on the CIFAR-10 test set reveal notable differences. FedAvg exhibits significant class confusion when learning new classes, likely due to shortcut learning. In contrast, STAMP, leveraging a prototypical coreset, effectively mitigates forgetting and maintains clearer class separation.}
\label{fig:tsne-proto}
\end{figure}

Figure~\ref{fig:tsne-proto} illustrates the effectiveness of prototype learning from a prototypical coreset. This figure highlights two key observations: (1) the inability of vanilla FL to effectively learn prototypes from hidden representations, and (2) the improved prototype learning capability achieved by STAMP. In the case of FedAvg, the model fails to acquire sufficiently representative features due to the limitations imposed by the single-pass data stream.

In contrast, STAMP demonstrates strong class discrimination as it progresses through tasks, which enhances its ability to learn prototypes from a compact coreset. This improvement stems from the coreset selection process, which is guided by class-specific criteria. As a result, it reduces inter-class confusion that could otherwise lead to inaccurate or misleading prototype representations.

\clearpage
\section{Privacy of STAMP}
FL \citep{2017-FL-FedAvg}, and FCL in particular, are vulnerable to various attacks such as data poisoning, model poisoning \citep{10423783}, backdoor attacks \citep{NEURIPS2023_d0c6bc64}, and gradient inversion attacks \citep{NEURIPS2024_9ff1577a, balunovic2022bayesian, dimitrov2022data}. Our proposed method does not introduce any additional privacy risks beyond those inherent to the standard FedAvg algorithm. Consequently, it is compatible with existing defense mechanisms developed for FedAvg, including secure aggregation \citep{} and noise injection prior to aggregation \citep{}.

Unlike several prior FCL approaches \citep{2023-FCL-TARGET, 2023-FCL-FedCIL} that require clients to share either locally trained generative models or perturbed private data, STAMP relies solely on gradient matching. It utilizes the global model weights and the uploaded local model updates, information already exchanged among clients in the standard FedAvg setting, thus avoiding the need for additional private data sharing, especially over open communication environments (e.g., 5G/6G wireless networks).

\section{Limitations and Future Works}\label{sec:limitations}
A primary limitation of our method lies in the sensitivity of gradient matching to the stability of task-wise and client-wise gradient trajectory approximation. Moreover, existing gradient matching approaches typically learn a single parameter set that adjusts the magnitude of task-specific gradients through a convex combination. Such approaches do not influence the direction of the gradients. Therefore, enhancing the stability of gradient trajectory approximation and improving gradient matching performance, particularly by extending the learnable parameter set to operate at the layer-wise or element-wise level, emerge as a promising direction for future research.

\end{document}